\documentclass[11pt]{article}

\usepackage[final]{acl}

\usepackage{times}
\usepackage{latexsym}
\usepackage[T1]{fontenc}

\usepackage[utf8]{inputenc}
\usepackage{float}
\usepackage{microtype}
\usepackage{inconsolata}
\usepackage{graphicx}
\usepackage{hyperref}
\usepackage{url}
\usepackage{CJKutf8}
\usepackage{booktabs}
\usepackage{textgreek}
\usepackage{subcaption} 
\usepackage{xspace}
\usepackage[many]{tcolorbox}
\usepackage{xcolor}
\usepackage{enumitem}
\usepackage{lscape}
\usepackage{wrapfig}
\usepackage[table]{xcolor}
\usepackage{titletoc}

\definecolor{proprietaryblue}{RGB}{222,235,247} 
\definecolor{opensourceblue}{RGB}{234,242,248}  
\definecolor{koreanblue}{RGB}{240,249,255}

\tcbset{colback=gray!5, colframe=gray!50, boxrule=0.5pt, arc=2pt}

\newcommand{\dataset}{HAERAE-Vision}

\title{What Users Leave Unsaid: Under-Specified Queries Limit \\ Vision-Language Models}

\author{
Dasol Choi\textsuperscript{1,2}\thanks{Equal contribution.} \quad
Guijin Son\textsuperscript{3}\footnotemark[1] \quad
Hanwool Lee\textsuperscript{1}\footnotemark[1] \quad
Minhyuk Kim\textsuperscript{4} \quad
Hyunwoo Ko\textsuperscript{3} \\
\textbf{Teabin Lim}\textsuperscript{5} \quad
\textbf{Eungyeol Ahn}\textsuperscript{5} \quad
\textbf{Jungwhan Kim}\textsuperscript{6} \quad
\textbf{Seunghyeok Hong}\textsuperscript{7}\thanks{Corresponding authors.} \quad
\textbf{Youngsook Song}\textsuperscript{8}\footnotemark[2]
\\[6pt]
\textsuperscript{1}AIM Intelligence \quad
\textsuperscript{2}Yonsei University \quad
\textsuperscript{3}OneLineAI \quad
\textsuperscript{4}Korea University \\
\textsuperscript{5}Doodlin Corp. \quad
\textsuperscript{6}NAVER Cloud \quad
\textsuperscript{7}Hankuk University of Foreign Studies \quad
\textsuperscript{8}Lablup Inc. \\
[5pt]
\raisebox{-0.2em}{\includegraphics[height=1em]{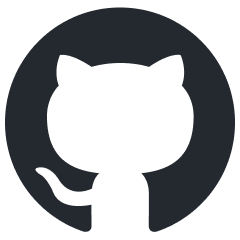}}~\href{https://github.com/HAE-RAE/HAE-RAE-VISION}{GitHub} \quad
\raisebox{-0.2em}{\includegraphics[height=1em]{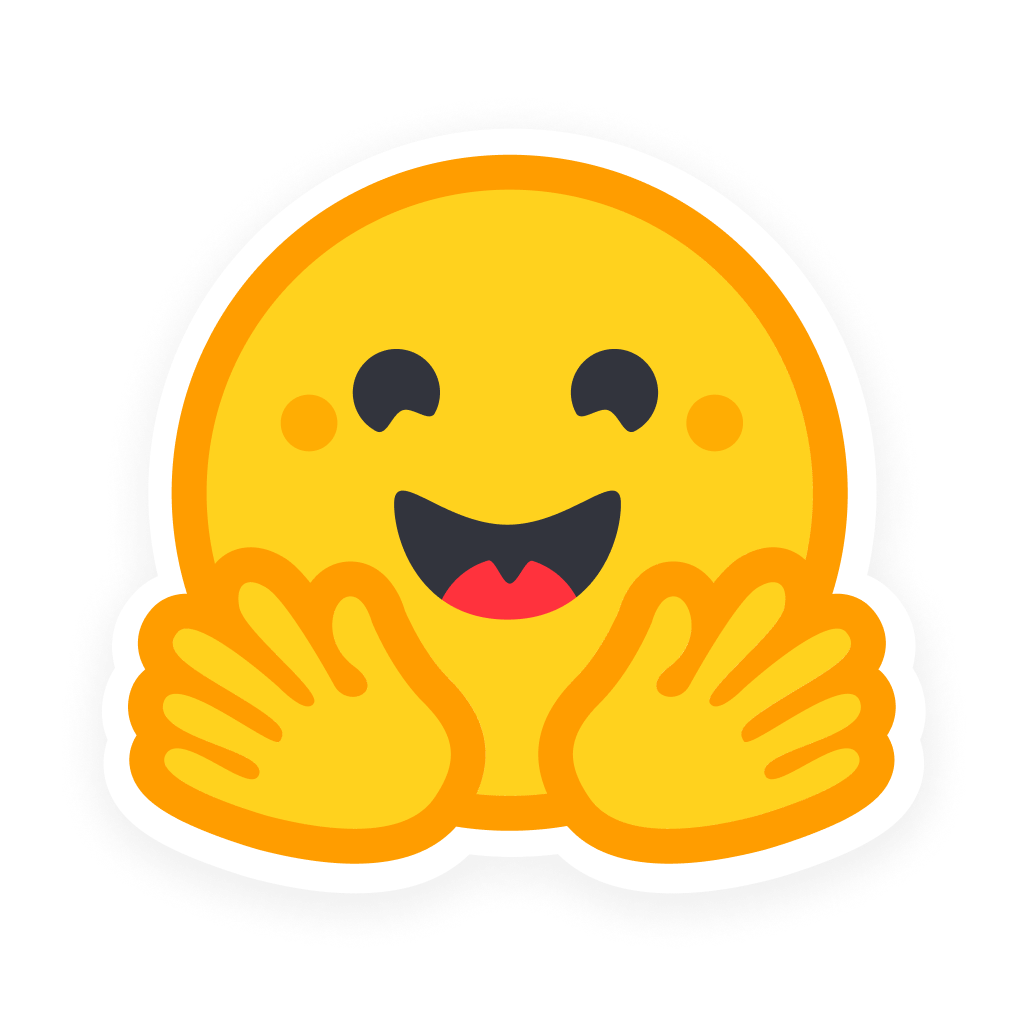}}~\href{https://huggingface.co/datasets/HAERAE-HUB/HAERAE-VISION}{HuggingFace} \quad
\raisebox{-0.2em}{\includegraphics[height=1em]{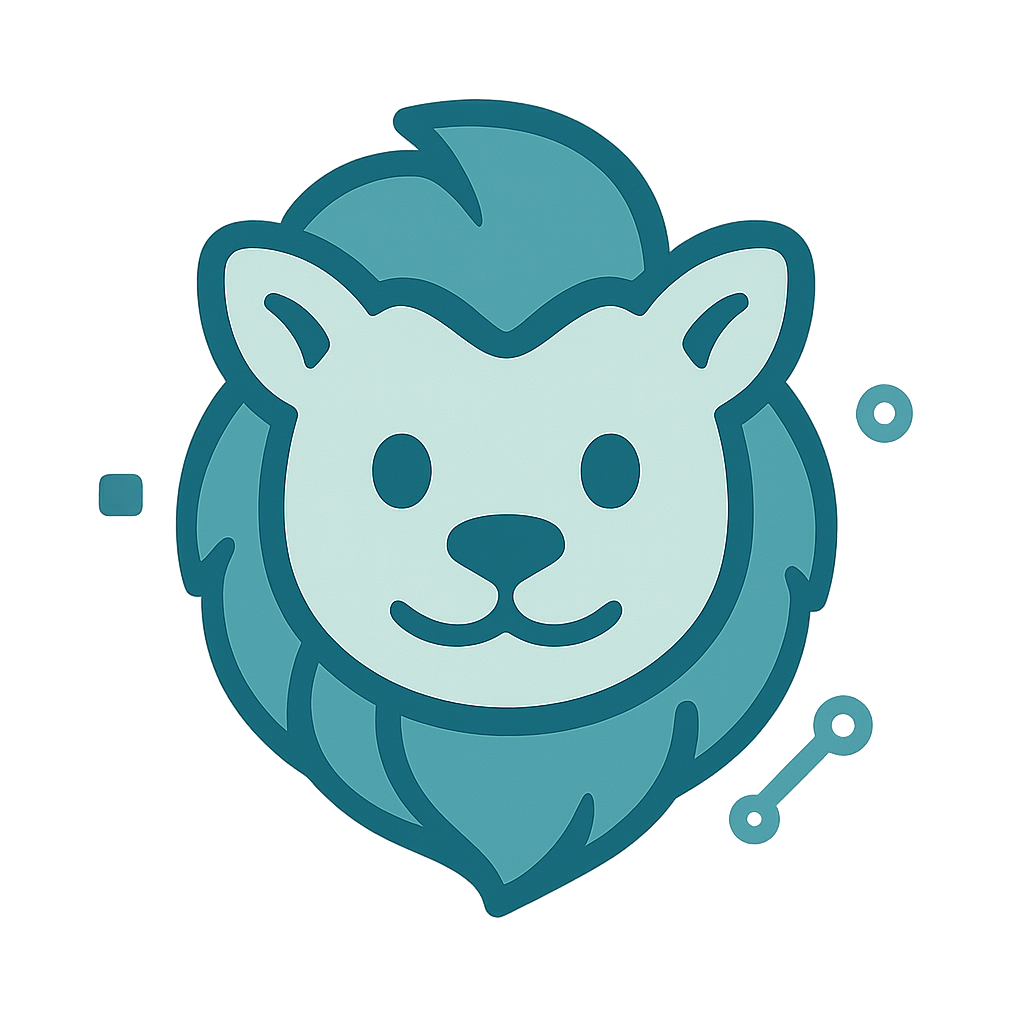}}~\href{https://board.haerae.world/}{Leaderboard}
\\[4pt]
\texttt{dasolchoi@yonsei.ac.kr, \; spthsrbwls123@yonsei.ac.kr}
}

\begin{document}

\maketitle

\begin{abstract}
Current vision-language benchmarks predominantly feature well-structured questions with clear, explicit prompts. However, real user queries are often informal and underspecified. Users naturally leave much unsaid, relying on images to convey context. We introduce HAERAE-Vision, a benchmark of 653 real-world visual questions from Korean online communities (0.76\% survival from 86K candidates), each paired with an explicit rewrite, yielding 1,306 query variants in total. Evaluating 45 VLMs, we find that even state-of-the-art models (GPT-5, Gemini 2.5 Pro) achieve under 50\% on the original queries. Crucially, query explicitation alone yields 8 to 22 point improvements, with smaller models benefiting most. 
We further show that even with web search, under-specified queries underperform explicit queries without search, revealing that current retrieval cannot compensate for what users leave unsaid.
Our findings demonstrate that a substantial portion of VLM difficulty stems from natural query under-specification instead of model capability, highlighting a critical gap between benchmark evaluation and real-world deployment.  
\end{abstract}

\section{Introduction}

When users ask visual questions, they rarely provide complete, well-structured queries. Instead, they write informally, omit context, and rely on images to convey what they leave unsaid. A user might ask ``How do I do this?'' alongside an image, expecting the responder to identify the problem, infer the relevant domain, and provide a step-by-step solution. This natural tendency toward under-specification poses a fundamental challenge for vision-language models (VLMs)~\citep{li2025questbench}, yet current benchmarks predominantly feature clean, explicit prompts failing to capture this phenomenon~\citep{KOFFVQA, VARCO-VISION}.

We introduce HAERAE-Vision, a benchmark constructed from authentic user queries in Korean online communities. Starting from 86,052 question-image pairs across nine platforms, we apply a six-stage filtering pipeline to yield 653 rigorously validated items (0.76\% survival rate). The resulting questions are ambiguous, informal, and under-specified, mirroring the noisy nature of authentic multimodal interactions. To isolate the effect of query under-specification, we additionally construct HAERAE-Vision-Explicit, a parallel dataset where each question is systematically rewritten to state the missing information explicitly.

Our experiments reveal that query explicitation alone yields up to 22 point improvements across models, with smaller models benefiting most dramatically. Even state-of-the-art models achieve under 50\% on original queries but surpass 55\% with explicitation 
(GPT-5: 48.0\%→57.6\%, Gemini 2.5 Pro: 48.5\%→56.7\%). Furthermore, we demonstrate that even with web search enabled, under-specified queries still underperform explicit queries without search. This reveals that current retrieval systems cannot compensate for what users leave unsaid, as models must first understand user intent before search becomes effective.

These findings challenge a common assumption in VLM evaluation: that benchmark difficulty reflects model capability limitations. We show that a substantial portion of difficulty stems instead from the natural under-specification of user queries, highlighting a critical gap between benchmark evaluation and real-world deployment.

\begin{figure*}[t]
\centering
\includegraphics[width=\linewidth]{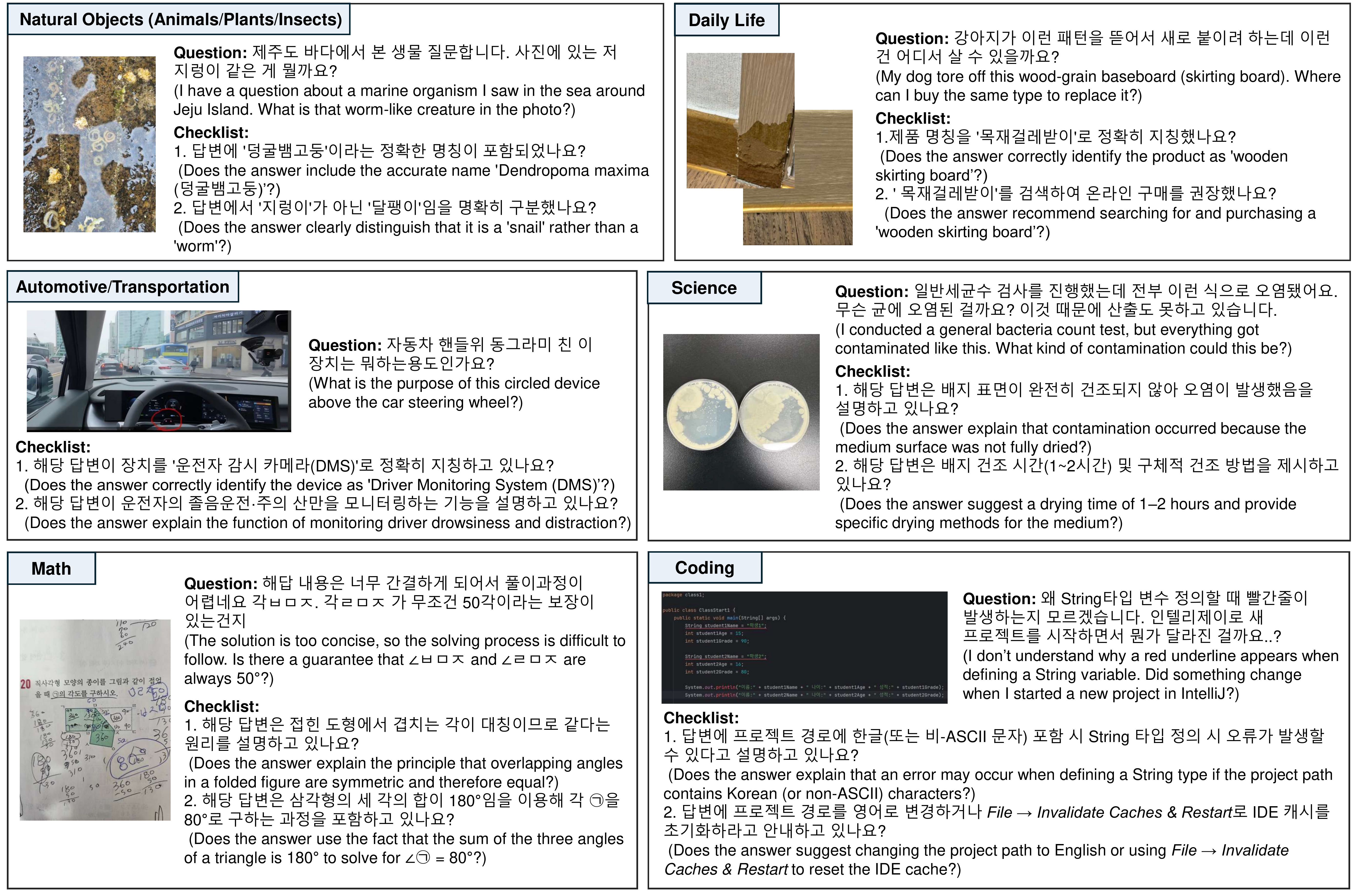}
\caption{\footnotesize \textbf{Representative examples from \dataset\ across six of the 13 domains.} Each example shows an under-specified Korean question with English translation, the corresponding image, and evaluation checklist criteria. Note the informal, context-dependent nature of the original queries.}
\label{fig:sample_items}
\end{figure*}

Our contributions are:

\begin{itemize}
\item \textbf{Real-world query benchmark:} HAERAE-Vision, comprising 
653 user-generated visual questions, filtered 
from 86K candidates (0.76\% survival), spanning 13 domains.
\item \textbf{Paired explicit rewrites:} A parallel dataset of clarified queries enabling 
controlled measurement of under-specification effects.
\item \textbf{Quantifying under-specification:} Empirical evidence 
that explicitation yields up to 22\% improvements, with smaller 
models benefiting most. This demonstrates that query ambiguity accounts for substantial VLM difficulty.
\end{itemize}

\begin{figure*}[t] 
\centering 
\includegraphics[width=0.97\textwidth]{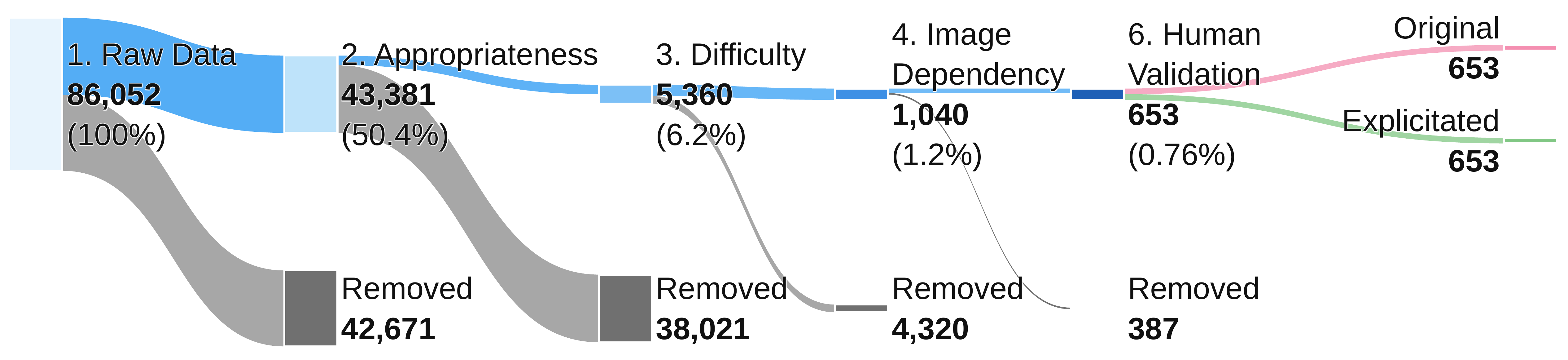} 
\caption{\footnotesize \textbf{Filtering pipeline showing data reduction at each stage.} Numbers indicate pipeline stages described in Section~\ref{subsec:pipeline}. The 0.76\% survival rate reflects rigorous quality control. Each validated question is paired with an explicitated rewrite, yielding 1,306 query variants.}
\label{fig:kvlm_funnel} 
\vspace{-3mm}
\end{figure*}

\section{HAERAE-Vision Benchmark}

We present \dataset, a benchmark constructed from authentic user queries, designed to capture the under-specified, informal nature of real-world visual questions. Our six-stage pipeline transforms large-scale, noisy community data into high-quality evaluation problems while preserving the natural characteristics of user queries.

\subsection{Dataset Construction Pipeline}
\label{subsec:pipeline}

Starting from 86,052 raw question-image pairs from nine Korean platforms spanning general Q\&A, gaming, science, and coding forums (see Appendix~\ref{app:platforms} for detailed platform descriptions), we obtain 653 high-quality problems (0.76\% survival rate). Figure~\ref{fig:kvlm_funnel} illustrates the filtering process.

\paragraph{Stage 1: Data Collection.} We collect (question, image, answer) triplets, prioritizing those with an accepted answer rewarded by the asker or with high online engagement (views, likes, comments), targeting questions the community finds valuable.

\paragraph{Stage 2: Appropriateness Filtering.} Each triplet is screened along three axes: (i) content safety (political/religious material, discrimination, adult content), (ii) objectivity (overly subjective or unverifiable prompts), and (iii) time-sensitiveness. GPT-4o is used for the automated filtering, flagging problematic items while allowing borderline cases to proceed to human validation. This removes 49.6\% of raw data (see Appendix~\ref{app:stage2_prompts}).

\paragraph{Stage 3: Difficulty Calibration.} Following prior benchmarks~\citep{zellers2019hellaswag, hendrycks2021measuring}, we remove questions that strong models solve trivially. Three models (GPT-4o, Gemini-1.5-Flash, Claude-3.5) are evaluated against community-provided human answers using semantic-overlap scoring. Items with an average score above 0.6 are removed, eliminating 87.6\% of the remaining items.

\paragraph{Stage 4: Image Dependency Verification.} To confirm that each question requires visual reasoning, we generate two responses per item using Gemini 2.0 Flash: one with the image and one without. Both responses are evaluated against the human reference, and items where the quality gap is below 1 point (on a 0-10 scale) are discarded as image-independent (see Appendix~\ref{app:stage4_prompt}).

\paragraph{Stage 5: Checklist Generation.} Each answer is converted into a structured checklist with 1 to 5 criteria using o4-mini. The model is instructed to define the minimal necessary elements for a response to be deemed correct, focusing on correctness, explanation quality, and reasoning steps rather than exhaustive coverage. This design enables partial-credit scoring and ensures reproducible, automated evaluation across models (see
Appendix~\ref{app:stage5_prompt}).

\paragraph{Stage 6: Human Validation.} Seven native Korean annotators conduct three-phase validation: (1) filtering based on image appropriateness, question clarity, and checklist validity, removing any item flagged by at least one annotator; (2) refinement of questions and LLM-generated checklists, where annotators rewrite unclear criteria and remove items not grounded in the original question–image pair; (3) final audit for category consolidation and consistency. This removes 37.2\% of remaining items, yielding 653 problems (see Appendix~\ref{app:guidelines}).


\begin{table}[t]
\centering
\renewcommand{\arraystretch}{0.8}
\fontsize{8}{8.5}\selectfont
\begin{tabular}{lcc}
\toprule
\textbf{Metric} & \textbf{Mean} & \textbf{Range} \\
\midrule
Q length (char) & 94.4 & 6--2{,}030 \\
Images per Q    & 1.3  & 1--6 \\
Checklist items & 3.3  & 1--5 \\
\midrule
\textbf{Category} & \textbf{\# Items} & \textbf{\%} \\
\midrule
Gaming & 91 & 13.9 \\
Entertainment/Arts & 50 & 7.7 \\
\midrule
Natural Objects & 92 & 14.1 \\
Science & 81 & 12.4 \\
Mathematics & 26 & 4.0 \\
\midrule
IT/Computer & 75 & 11.5 \\
Coding/Development & 45 & 6.9 \\
Machinery & 22 & 3.4 \\
\midrule
Daily Life & 51 & 7.8 \\
Business/Economics & 37 & 5.7 \\
Transportation & 35 & 5.4 \\
Shopping/Consumer & 27 & 4.1 \\
Health/Medical & 21 & 3.2 \\
\midrule
\textbf{Total} & \textbf{653} & \textbf{100.0} \\
\bottomrule
\end{tabular}
\caption{\footnotesize \textbf{Overview of \dataset.} Statistics of question length, number of images, and checklist items, highlighting the diversity and multimodal nature of \dataset.}
\label{tab:overview_combo}
\end{table}

\subsection{Dataset Statistics}

Our final benchmark contains 653 problems with an average of 3.3 checklist items and 1.3 images per question. Table~\ref{tab:overview_combo} presents the distribution across 13 categories, where Natural Objects and Gaming are the most represented. The survival rate per platform varies significantly (0.2\% to 14.4\%), showing distinct community characteristics (see Appendix~\ref{app:platform_stats} for detailed breakdown).

\begin{figure*}[t]
\centering
\small
\renewcommand{\arraystretch}{0.8}
\setlength{\tabcolsep}{4pt}
\begin{tabular}{@{}m{0.18\textwidth} m{0.36\textwidth} m{0.4\textwidth}@{}}
\toprule
\textbf{Image} & \textbf{Original} & \textbf{Explicitated} \\
\midrule
\includegraphics[width=0.08\textwidth]{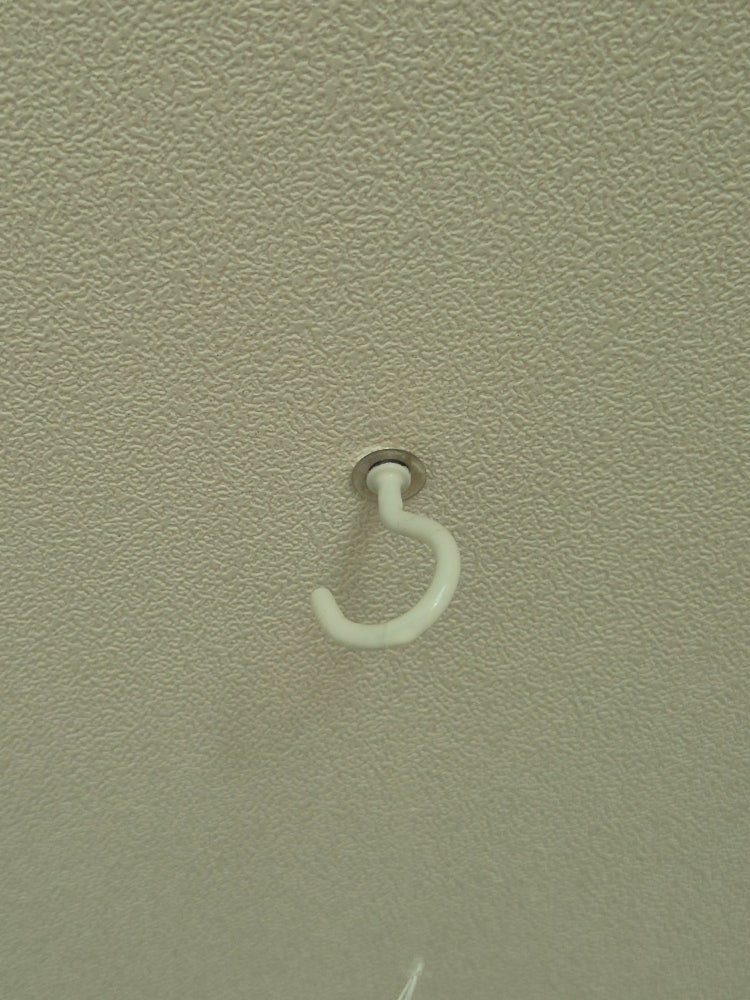}
\includegraphics[width=0.08\textwidth]{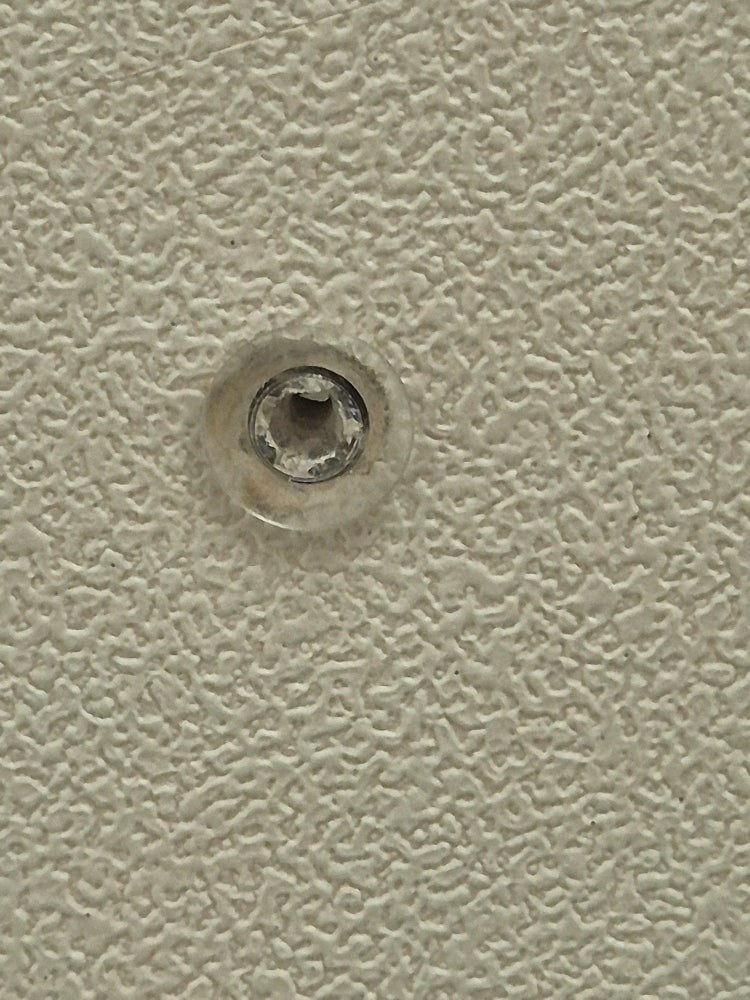} & 
{\scriptsize \begin{CJK}{UTF8}{mj}이거는 어떻게 빼는걸까요? 저 고리를 빼고나니 저렇게 남았는데 저부분은 어떻게 빼야하나요?\end{CJK}} \newline {\scriptsize (How do I remove this? After removing the hook, this part remains—how do I take it out?)} & 
{\scriptsize \begin{CJK}{UTF8}{mj}천장에 설치된 흰색 고리형 행거를 제거한 후 남은 금속 부속품을 완전히 분리하려면 어떻게 해야 하나요?\end{CJK}} \newline {\scriptsize (How do I completely remove the metal fitting left after detaching the white ceiling hook hanger?)} \\
\midrule
\includegraphics[width=0.17\textwidth]{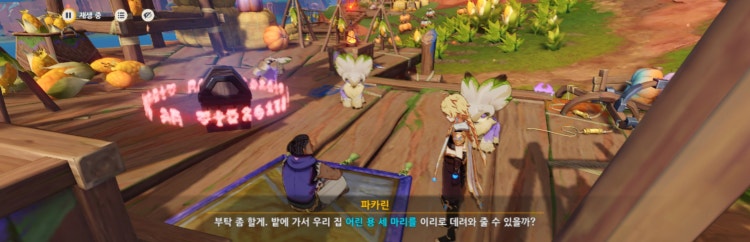} & 
{\scriptsize \begin{CJK}{UTF8}{mj}어린용 저 3마리 말고 더 있나요?\end{CJK}} \newline {\scriptsize (Are there more besides those 3 baby dragons?)} & 
{\scriptsize \begin{CJK}{UTF8}{mj}게임 '원신'에서 파카틴 NPC가 의뢰하는 임무 중 등장하는 이 어린 용 세 마리 외에 추가로 찾아야 하는 용이 더 있나요?\end{CJK}} \newline {\scriptsize (In Genshin Impact, are there additional dragons to find beyond the three baby dragons in Parkatin's quest?)} \\
\midrule
\fbox{\includegraphics[width=0.17\textwidth]{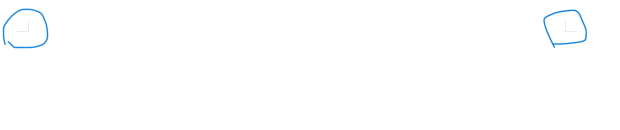}} & 
{\scriptsize \begin{CJK}{UTF8}{mj}한글 머리말 경계선 없애는 법. 동그라미 친 부분 없앨 수 있나요?\end{CJK}} \newline {\scriptsize (How to remove header border in Hangul. Can I remove the circled part?)} & 
{\scriptsize \begin{CJK}{UTF8}{mj}한글 문서에서 머리말 구역 상단에 표시되는 여백 경계선을 제거하려면 어떻게 해야 하나요?\end{CJK}} \newline {\scriptsize (How do I remove the margin border line shown at the top of the header area in Hangul word processor?)} \\
\bottomrule
\end{tabular}
\caption{\footnotesize \textbf{Examples of query explicitation across three domains (Daily Life, Gaming, IT/Software).} Original queries contain vague references that depend on images. Explicitated versions include background information to clarify the user request.}
\label{fig:explicitation_examples}
\vspace{-3mm}
\end{figure*}

\subsection{HAERAE-Vision-Explicit}

To isolate the effect of query under-specification, we construct a parallel dataset where each question is rewritten to explicitly state the missing information while preserving the original intent. Figure~\ref{fig:explicitation_examples} illustrates the transformation from under-specified to explicit queries across different domains.

We use GPT-5.1 with web search to rewrite each question following strict guidelines (Appendix~\ref{app:explicitation_prompt}): (1) preserve the original intent and scope without broadening or narrowing, (2) make implicit context explicit by specifying domains, entities, and concrete references, (3) replace vague references such as ``this,'' ``that,'' or ``here,'' (4) incorporate visual information from the image into the question, and (5) use web search only to verify proper nouns (e.g., game titles, product names) implied by the original question. 
Each rewritten question then undergoes human validation. Three annotators reviewed all 653 explicitated questions against their corresponding images, verifying factual accuracy, correcting hallucinated details through additional search, and adjusting specificity by removing overly specific terms or adding missing context where necessary. This process yields 653 explicitated questions paired with the original under-specified versions.

\begin{figure*}[ht]
\centering
\includegraphics[width=0.98\textwidth]{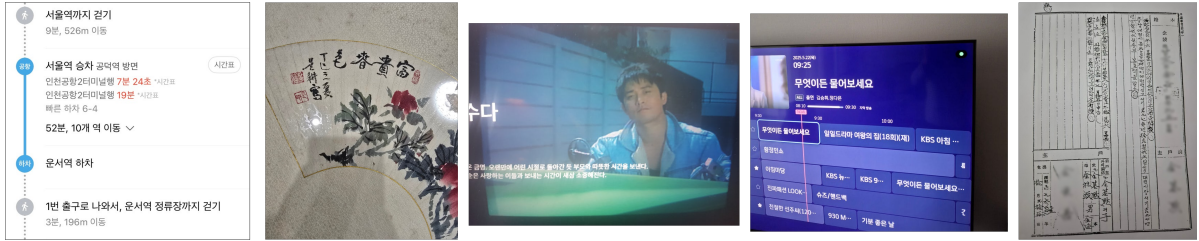}
\caption{\footnotesize \textbf{Examples highlighting the cultural specificity of \dataset:} (a) Seoul subway interface, (b) traditional painting with calligraphy, (c) Korean drama scene requiring celebrity recognition, (d) TV channel settings, (e) historical family registry. Such culturally grounded items require knowledge rarely represented in English-centric datasets.}
\label{fig:korean_examples}
\vspace{-3mm}
\end{figure*}

\subsection{Korean Cultural Grounding}

We consider an item \textit{culturally grounded} if it 
requires knowledge of Korean institutions, services, policies, 
local brands or products, or 
Korean-language UI and text conventions; items solvable through 
globally shared knowledge are marked non-cultural.
Under this criterion, 23.7\% of items require distinctively Korean 
cultural knowledge, including local interfaces (Seoul Metro signage, 
Naver SmartPlace), region-specific objects (winter road sandbags), 
or Korean media (drama actors, traditional calligraphy). These items 
are rarely represented in English-centric training corpora.
Figure~\ref{fig:korean_examples} shows representative examples.

\section{Evaluation Framework}

\subsection{Checklist-based Assessment}
To mitigate the subjectivity of single-label scoring and the noise inherent in raw web text, our methodology centers on detailed checklists that decompose complex answers into specific criteria. Supported by recent findings that instance-specific rubrics align better with human judgments \citep{kim2024biggen}, each problem includes 1--5 evaluation points assessing different reasoning aspects.
This checklist approach provides several advantages over traditional methods:
(1) Fine-grained assessment of partial understanding,
(2) Reduced subjectivity through explicit criteria,
(3) Diagnostic capability for pinpointing model weaknesses, and
(4) Scalability for automated evaluation.

\subsection{LLM Judge Protocol}
GPT-5-Mini is instructed to act as the primary judge, following a structured prompt that enforces consistent scoring across all problems (Appendix~\ref{app:judge_prompt}). Each checklist item is scored on a three-level scale: \emph{met} (1.0), \emph{partially met} (0.5), or \emph{not met} (0.0), based solely on explicit evidence found in the model's response. Each score is accompanied by supporting evidence and justification, where the evidence is a single line directly extracted from the response and the justification is a short rationale clarifying the decision. The model outputs a structured report containing evidence blocks and fractional totals (e.g., \texttt{3.5/5} when one item is partially and three are fully satisfied out of five). The overall score is computed as the average of instance-level means, where each instance has $m_i$ checklist items with item scores $r_{ij} \in \{0, 0.5, 1\}$:
\[
S_{\text{final}} = \frac{1}{N} \sum_{i=1}^{N} \left( \frac{1}{m_i} \sum_{j=1}^{m_i} r_{ij} \right),
\]
ensuring comparability across problems with differing checklist lengths.

\section{Experimental Setup}

\subsection{Model Selection}

We evaluate 45 VLMs covering a broad range of families and scale. \textbf{Proprietary models.} This group includes OpenAI’s GPT-5 series (\texttt{GPT-5}, \texttt{GPT-5-Mini}, \texttt{GPT-5-Nano})~\citep{openai2025gpt5systemcard}, 
Google’s Gemini (2.5-Pro, 2.5-Flash, 2.5-Flash-Lite)~\citep{deepmind2025gemini25procard}, 
and proprietary systems such as 
\texttt{Perplexity-Sonar-Pro}~\citep{perplexity2025sonarpro}, 
\texttt{xAI-Grok-4}~\citep{xai2025grok4card}, Mistral (\texttt{Medium-3.1}, \texttt{Small-24B}) and Pixtral (\texttt{Large}, \texttt{12B})~\citep{mistral2024pixtralLarge2411,pixtral12b_2024}. \textbf{Open-source models.} We evaluate
\texttt{Gemma-3} (27B, 12B, 4B)~\citep{gemma3_techreport_2025}, 
\texttt{Qwen2.5-VL} (72B, 7B, 3B)~\citep{qwen25vl_2025}, 
\texttt{Qwen3-VL} (235B-A22B, 32B, 30B-A3B, 8B, 4B, 2B; each in \textit{Instruct} and \textit{Thinking} variants)~\citep{qwen3},
\texttt{Skywork-R1V3-38B}~\citep{skywork_r1v3_2025}, 
\texttt{InternVL3.5} (38B--1B)~\citep{internvl35_2025}, 
and \texttt{AIDC-AI-Ovis2} (34B--1B)~\citep{ovis25_techreport_2025}. \textbf{Korean models.} Finally, we include Korean-specific models, including 
\texttt{VARCO-VISION-2.0} (14B, 1.7B)~\citep{varco_vision_20_2025} 
and \texttt{HyperCLOVA-3B}~\citep{hyperclovax_2024}.

\subsection{Implementation Details}

We used \texttt{temperature=0.6} (1.0 for GPT-5 due to provider constraints), \texttt{top\_p=0.95}, and \texttt{max\_tokens=4096} across all models. Each instance was evaluated three times and averaged.


\begin{table*}[ht]
\centering
\renewcommand{\arraystretch}{0.85}
\fontsize{9}{9.5}\selectfont
\begin{tabular}{@{}lccccc@{}}
\toprule
Model & Entertainment & Scientific & Technical & Daily Life & Overall \\
\midrule
\multicolumn{6}{l}{\textit{Proprietary Models}} \\
Gemini 2.5 Pro & $\mathbf{40.52_{0.61}}$ & $\mathbf{51.44_{0.40}}$ & $53.89_{0.79}$ & $52.64_{0.93}$ & $\mathbf{48.54_{0.11}}$ \\
GPT-5 & $33.07_{0.87}$ & $48.14_{0.96}$ & $\mathbf{55.71_{0.84}}$ & $\mathbf{55.98_{0.75}}$ & $48.01_{0.19}$ \\
GPT-5 Mini & $27.38_{0.81}$ & $50.62_{0.93}$ & $51.88_{0.74}$ & $51.31_{1.32}$ & $45.21_{0.70}$ \\
Perplexity Sonar-Pro & $32.84_{0.76}$ & $47.98_{0.59}$ & $47.17_{1.23}$ & $49.64_{0.64}$ & $44.28_{0.48}$ \\
Gemini 2.5 Flash & $29.31_{1.09}$ & $45.04_{0.98}$ & $44.05_{0.53}$ & $48.72_{1.38}$ & $41.05_{0.79}$ \\
Grok-4 & $26.88_{0.67}$ & $31.03_{0.64}$ & $44.18_{0.80}$ & $39.67_{0.55}$ & $36.08_{0.30}$ \\
Gemini 2.5 Flash-Lite & $18.39_{0.59}$ & $38.17_{1.47}$ & $32.74_{0.84}$ & $35.47_{0.92}$ & $30.29_{0.24}$ \\
GPT-5 Nano & $11.64_{0.53}$ & $20.10_{1.24}$ & $27.15_{1.36}$ & $29.68_{0.54}$ & $21.22_{0.26}$ \\
\midrule
\multicolumn{6}{l}{\textit{Open Source Models}} \\
Skywork-R1V3-38B & $15.03_{0.73}$ & $35.31_{0.88}$ & $30.22_{0.49}$ & $33.75_{0.72}$ & $27.76_{0.34}$ \\
Mistral Medium 3.1 & $13.74_{0.80}$ & $30.77_{0.86}$ & $28.87_{0.67}$ & $28.78_{1.01}$ & $24.86_{0.56}$ \\
Gemma-3 27B & $11.59_{0.58}$ & $25.80_{0.61}$ & $22.28_{1.04}$ & $30.85_{0.61}$ & $22.53_{0.16}$ \\
Qwen2.5-VL-72B & $10.89_{0.66}$ & $26.71_{1.49}$ & $21.60_{0.53}$ & $25.61_{0.52}$ & $20.58_{0.46}$ \\
Pixtral Large & $11.43_{0.82}$ & $21.79_{0.50}$ & $21.77_{0.38}$ & $25.65_{0.91}$ & $20.10_{0.24}$ \\
InternVL3.5-38B & $8.81_{0.46}$ & $23.25_{0.61}$ & $17.92_{0.73}$ & $23.36_{0.78}$ & $18.01_{0.22}$ \\
Ovis2-34B & $9.52_{0.47}$ & $21.88_{0.55}$ & $21.00_{0.51}$ & $24.82_{0.58}$ & $18.50_{0.02}$ \\
Mistral Small 24B & $6.46_{0.29}$ & $10.18_{0.45}$ & $13.30_{0.66}$ & $16.20_{0.66}$ & $11.20_{0.01}$ \\
\midrule
\multicolumn{6}{l}{\textit{Korean-specialized Models}} \\
VARCO-VISION 2.0 (14B) & $7.87_{0.80}$ & $16.56_{0.65}$ & $16.88_{0.57}$ & $22.13_{0.88}$ & $15.55_{0.29}$ \\
HyperCLOVA X-SEED-3B & $6.25_{0.25}$ & $14.87_{0.51}$ & $11.99_{0.50}$ & $17.93_{0.73}$ & $12.66_{0.10}$ \\
\bottomrule
\end{tabular}%
\caption{\footnotesize \textbf{Performance of 18 models averaged by category.} For model families with multiple sizes, only the largest variant is shown.  Full results across all model sizes and detailed 13-category breakdowns are in Appendix~\ref{tab:detailed_all_categories}. 
All scores are reported as mean$_{SE}$, where SE is the standard error over 3 independent runs (n=3). The highest-scoring model is highlighted in \textbf{bold}.}
\label{tab:category_results_std}
\end{table*}

\section{Results}

\subsection{Overall Performance}

Table~\ref{tab:category_results_std} summarizes the performance of 18 VLMs across four categories (full results are provided in Appendix~\ref{app:full_results}). 
Even the best-performing models—Gemini 2.5 Pro (48.5\%) and GPT-5 (48.0\%)—fall short of 50\% accuracy, 
highlighting that authentic, culturally grounded multimodal queries remain far from solved.
Proprietary systems consistently outperform open-weight counterparts, with the strongest open-weight models 
(Skywork-R1V3-38B: 27.8\%, Qwen2.5-VL-72B: 20.6\%) reaching roughly half the accuracy of top proprietary models. 
Neither search-augmented models (Perplexity Sonar-Pro: 44.3\%) nor reasoning-specialized models (Skywork-R1V3) 
achieve notable gains, suggesting that solving would require capabilities beyond current 
retrieval-augmented or reasoning-optimized paradigms.

Korean-specialized models also struggled to achieve competitive results (VARCO-VISION 2.0 14B: 15.6\%, HyperCLOVA X-SEED-3B: 12.7\%), indicating that dedicated local models have yet to demonstrate clear advantages on this benchmark. See Appendix~\ref{app:scale_domain} for a domain-level analysis.

\subsection{Effect of Query Explicitation}

\begin{figure}[t]
\centering
\includegraphics[width=0.48\textwidth]{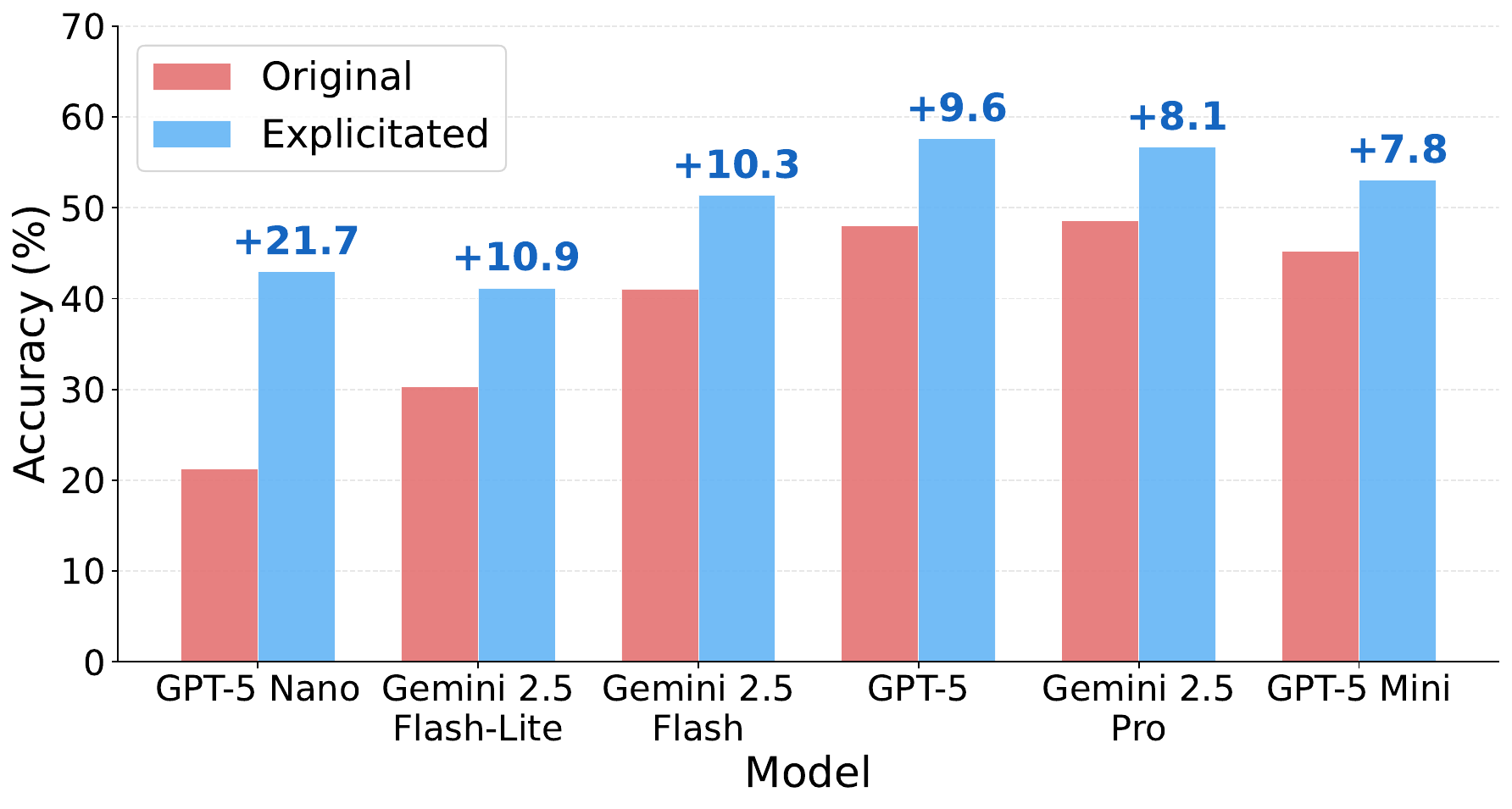}
\caption{\footnotesize \textbf{Effect of query explicitation on model performance.} Models are sorted by improvement magnitude. Smaller models benefit most from explicitation, with GPT-5 Nano showing +21.7 points improvement. All results averaged over 3 runs.}
\label{fig:explicitation_effect}
\end{figure}

Figure~\ref{fig:explicitation_effect} shows the effect of query explicitation on model performance. Across all six models, explicitation yields substantial improvements of 7.8 to 21.7 points.
Smaller models benefit most from explicitation: GPT-5-Nano improves by 21.7 points (21.2 → 43.0), more than doubling its performance, while larger models like GPT-5 and Gemini 2.5 Pro show gains of 9.6 and 8.1 points respectively. This pattern suggests that under-specified queries disproportionately disadvantage smaller models, which may lack the capacity to infer implicit context from images alone.
Even with explicitation, the best-performing model (GPT-5) achieves only 57.6\%, indicating that query under-specification accounts for a substantial portion, but not all, of the difficulty in \dataset{}. Our error analysis (Section~\ref{sec:analysis}) reveals that the remaining challenges stem primarily from cultural knowledge gaps.

\subsection{Effect of Web Search}

To isolate the contributions of query explicitation and retrieval augmentation, we evaluated GPT-5 and GPT-5-Mini across all four conditions: original and explicitated queries, each with and without web search. We use the official OpenAI search API \cite{openai_websearch}.

\begin{table}[t]
\centering
\renewcommand{\arraystretch}{0.9}
\small
\begin{tabular}{lcccc}
\toprule
\textbf{Model} & \textbf{Orig} & \textbf{Orig+S} & \textbf{Expl} & \textbf{Expl+S} \\
\midrule
GPT-5 & 48.01 & 55.58 & 57.57 & 59.72 \\
GPT-5-Mini & 45.21 & 51.08 & 53.04 & 56.69 \\
\midrule
\multicolumn{5}{l}{\textit{Δ from Original (no search)}} \\
GPT-5 & -- & +7.57 & +9.56 & +11.71 \\
GPT-5-Mini & -- & +5.87 & +7.83 & +11.48 \\
\bottomrule
\end{tabular}
\caption{\footnotesize \textbf{Effect of web search and query explicitation.} Scores reported as mean over 3 runs. Original+Search still underperforms Explicit alone, indicating retrieval cannot compensate for under-specification.}
\label{tab:search_results}
\end{table}

As shown in Table~\ref{tab:search_results}, web search yields moderate improvements for original queries (GPT-5: +7.57; GPT-5-Mini: +5.87), but these gains are smaller than those obtained through explicitation alone (+9.56 and +7.83, respectively). Notably, original queries augmented with search still underperform explicit queries without search (GPT-5: 55.58 vs. 57.57; GPT-5-Mini: 51.08 vs. 53.04). This indicates that retrieval cannot compensate for under-specified queries; models must first infer user intent for search to be effective. We observe a recurring failure mode in which models rely on textual cues during search while failing to ground visual features, suggesting that current web search integration operates at a largely surface level and is not deeply leveraged by GPT-5. The highest performance is achieved when explicitation and search are combined (GPT-5: 59.72; GPT-5-Mini: 56.69), demonstrating additive benefits. However, the marginal improvement from adding search to explicit queries (+2.15 and +3.65) is smaller than when added to original queries, implying that explicitation already supplies much of the contextual information that search would otherwise retrieve.

\subsection{Cross-Lingual Validation}
\label{sec:cross_lingual}

To test whether the explicitation effect generalizes beyond Korean, we conduct a pilot study in English. We collect approximately 3,000 image-containing Q\&A pairs from 12 Stack Exchange communities spanning 9 of our 13 categories and apply the same six-stage pipeline (Section~\ref{subsec:pipeline}). After filtering, 627 samples survive; we randomly select 100 samples stratified by category for evaluation (see Appendix~\ref{app:english_pilot} for full construction details and per-domain results).

\begin{table}[t]
\centering
\renewcommand{\arraystretch}{0.9}
\small
\begin{tabular}{lccc}
\toprule
\textbf{Model} & \textbf{Original} & \textbf{Explicit} & \textbf{$\Delta$} \\
\midrule
GPT-5 & 60.8 & 65.6 & +4.8 \\
GPT-5-Mini & 53.3 & 59.9 & +6.6 \\
Gemini 2.5 Pro & 51.3 & 57.3 & +6.0 \\
GPT-5-Nano & 44.4 & 47.6 & +3.2 \\
\bottomrule
\end{tabular}
\caption{\footnotesize \textbf{English pilot: effect of query explicitation.} All four models show consistent gains, confirming that underspecification effects are cross-lingual. Deltas are smaller than in Korean (+3.2--6.6 vs.\ +7.8--21.7).}
\label{tab:english_pilot}
\end{table}

As shown in Table~\ref{tab:english_pilot}, all four models show consistent explicitation gains (+3.2 to +6.6 points), confirming that the effect of query underspecification is not limited to Korean. However, English deltas are consistently smaller than their Korean counterparts (+7.8 to +21.7). Notably, GPT-5-Nano scores 44.4\% on English original queries, more than double its Korean score (21.2\%), suggesting that smaller models can better compensate for underspecification in high-resource languages.

Root-cause analysis on remaining errors after explicitation reveals why the gap is smaller in English: cultural knowledge accounts for only 2.7\% of English errors versus 19.0\% in Korean, with general reasoning comprising 96.7\% of English failures. This confirms that the larger Korean explicitation gap is driven by the interaction between surface-level underspecification and culturally grounded assumptions that are underrepresented in English-centric training corpora. Once explicitation resolves surface ambiguity, cultural knowledge remains as a persistent source of difficulty in Korean but not in English.

\section{Additional Analysis on Explicitation}
\label{sec:analysis}

To understand why explicitation improves performance, we analyzed error patterns across original and explicitated conditions. We collected 3,164 (original) and 2,834 (explicitated) error cases where models scored below 1.0, spanning six models (GPT-5, GPT-5-Mini, GPT-5-Nano, Gemini 2.5 Pro, Gemini 2.5 Flash, Gemini 2.5 Flash-Lite).
Each error was annotated by an LLM judge (Claude 3.5 Sonnet) along two dimensions: (1) \textit{failure category}---how the error manifests (Table~\ref{tab:error_taxonomy_brief}); and (2) \textit{root cause}---why the error occurs. Failure categories are multi-label (1--3 per error), while root causes are single-label; the two dimensions are orthogonal, so their percentages are not directly comparable. The full annotation prompt and category definitions are provided in Appendix~\ref{app:error_annotation}.

\begin{table}[t]
\centering
\fontsize{8}{8.5}\selectfont
\begin{tabular}{ll}
\toprule
\textbf{Failure Category} & \textbf{Description} \\
\midrule
Lack of explicitness & Missing checklist-required facts \\
Procedural reasoning & Failed multi-step execution \\
Object recognition & Misidentified visual entities \\
Cultural mismatch & Misunderstood Korean conventions \\
Visual-text grounding & Wrong image region referenced \\
Spatial reasoning & Incorrect spatial relations \\
\midrule
\textbf{Root Cause} & \\
\midrule
General reasoning & Logic/inference failure \\
Cultural knowledge & Missing Korean-specific knowledge \\
Language & Korean language misunderstanding \\
\bottomrule
\end{tabular}
\caption{\footnotesize \textbf{Error annotation taxonomy (abbreviated).}}
\label{tab:error_taxonomy_brief}
\end{table}

\begin{table}[t]
\centering
\small
\begin{tabular}{lccc}
\toprule
\textbf{Failure Category} & \textbf{Orig} & \textbf{Expl} & \textbf{Δ} \\
\midrule
Lack of explicitness & 84.3\% & 69.7\% & -14.6 \\
Procedural reasoning & 66.6\% & 64.3\% & -2.3 \\
Object recognition & 20.6\% & 18.5\% & -2.1 \\
Cultural concept mismatch & 13.1\% & 22.5\% & +9.4 \\
Visual-text grounding & 5.2\% & 16.6\% & +11.4 \\
\bottomrule
\end{tabular}
\caption{\footnotesize \textbf{Failure category shifts from original to explicitated queries.}}
\label{tab:error_shift}
\end{table}

\subsection{What Explicitation Fixes}

Table~\ref{tab:error_shift} shows the key shifts. The most striking change is the reduction in \textit{lack of explicitness} failures, which drop from 84.3\% to 69.7\% (-14.6pp), directly confirming that explicitation addresses surface-level ambiguity. 
Smaller models show the largest reductions in error cases after explicitation (GPT-5-Nano: -83 cases, +12.7pp perfect rate) compared to larger models (GPT-5-Mini: -40 cases, +6.1pp), confirming that under-specification disproportionately impacts smaller models.

\begin{figure}[ht]
\hspace{-0.45cm}
\centering
\includegraphics[width=0.48\textwidth]{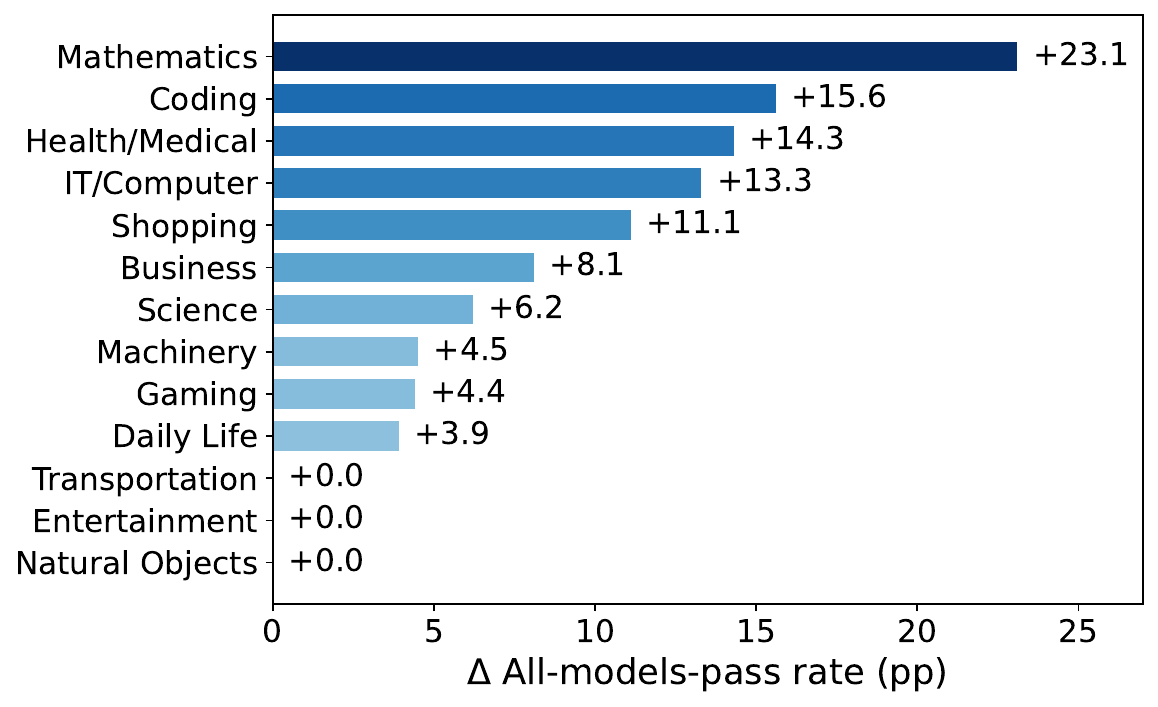}
\caption{\footnotesize \textbf{Category-level explicitation effects.} Categories like Mathematics and Coding show large gains, while Entertainment and Natural Objects remain difficult even after clarification, with failures shifting toward cultural knowledge and visual grounding.}
\label{fig:category_effect}
\end{figure}

Category-level analysis (Figure~\ref{fig:category_effect}) reveals 
that explicitation yields the largest gains in Mathematics, Science, Coding, and Shopping—categories where failures primarily stemmed from under-specified problem descriptions. In contrast, Natural Objects and Entertainment remain challenging even after clarification (all-models-pass rate: 0\% in both conditions), with failures shifting toward visual-text grounding and cultural knowledge gaps.

Notably, visual-text grounding (VTG) errors increase from 5.2\% to 16.6\% after explicitation. However, tracking individual error cases reveals that this reflects an \textit{unmasking} effect rather than a trade-off: 87\% of VTG errors in the explicitated condition were already errors under original queries but classified under other categories (primarily lack of explicitness). Explicitation forces models to engage with specific visual regions, exposing latent grounding failures previously obscured by surface-level ambiguity (see Appendix~\ref{app:vtg_unmasking} for detailed analysis).

\subsection{Why Retrieval Alone Is Insufficient}

Earlier, our results have shown that original queries with search (55.6) underperform explicit queries without search (57.6). This reveals a fundamental limitation: retrieval cannot compensate for query under-specification. Under-specified queries like ``\begin{CJK}{UTF8}{mj}이거 어떻게 해요?\end{CJK}'' (How do I do this?) contain no searchable keywords. Since the critical context is embedded solely within the visual modality, current text-based search engines fail to bridge the modality gap without explicit textual grounding. Even when models attempt searches, they lack the specific terms (product names, game titles, error codes) needed to retrieve useful results.
In contrast, explicitated queries contain concrete references (e.g., ``\begin{CJK}{UTF8}{mj}천장에 설치된 흰색 고리형 행거\end{CJK}'' (white ring-shaped hanger installed on the ceiling)) that enable targeted retrieval. The best performance is achieved when both are combined (59.7), but the key finding is that search on under-specified queries cannot match explicitation alone; models must first understand what to search for.

\subsection{Cultural Knowledge Gaps}

After explicitation, what errors remain? Analyzing root causes reveals a shift toward cultural knowledge gaps (Table~\ref{tab:root_cause}).
The increase in cultural knowledge attribution (+6.4pp) suggests that once query ambiguity is resolved, the dominant remaining challenge is Korea-specific knowledge. For example, when shown orange bags along a rural road, models identified them as ``road safety markers'' or ``wasp traps,'' missing that these are winter snow preparation sandbags, something all native Korean drivers would have known. Similarly, all SOTA models misidentified a Korean folder phone (SKY IM-100) as global brands like Sony or Nokia. Finally, the negligible language error rate (<1.5\%) confirms that Korean proficiency is no longer a hurdle for global models, but cultural content is.

\begin{table}[t]
\centering
\small
\renewcommand{\arraystretch}{0.85}
\begin{tabular}{lccc}
\toprule
\textbf{Root Cause} & \textbf{Orig} & \textbf{Expl} & \textbf{Δ} \\
\midrule
General reasoning & 86.6\% & 79.8\% & -6.9 \\
Cultural knowledge & 12.7\% & 19.0\% & +6.4 \\
Language & 0.7\% & 1.2\% & +0.5 \\
\bottomrule
\end{tabular}
\caption{\footnotesize \textbf{Root cause distribution.} After explicitation, cultural knowledge becomes more prominent as surface-level ambiguity is resolved.}
\label{tab:root_cause}
\end{table}

\begin{table}[t]
\centering
\fontsize{7.6}{9}\selectfont
\setlength{\tabcolsep}{2pt}
\begin{tabular}{lcccc}
\toprule
& \textbf{GPT-5-mini} & \textbf{GPT-5} & \textbf{Gem-2.5-Pro} & \textbf{Gem-2.5-Flash} \\
\midrule
GPT-5-mini       & --  & 0.87 & 0.90 & 0.90 \\
GPT-5            & 0.87 & --  & 0.90 & 0.86 \\
Gem-2.5-Pro   & 0.90 & 0.90 & --  & 0.89 \\
Gem-2.5-Flash & 0.90 & 0.86 & 0.89 & --  \\
\midrule
\addlinespace[2pt]
\multicolumn{5}{l}{\textbf{Krippendorff's \(\alpha = 0.867\)}} \\
\bottomrule
\end{tabular}
\caption{\footnotesize \textbf{Pairwise Pearson correlations among four LLM judges.} Spearman correlations range 0.87–0.90. Krippendorff's $\alpha = 0.867$ indicates substantial agreement.}
\label{tab:interjudge}
\end{table}

\section{Reliability of LLM-as-a-Judge}
It is widely known that LLM-Judges may be prone to biases~\citep{son2024llm}. Accordingly, to ensure the credibility of our evaluation, we assess the inter-judge agreement among four LLM judges (GPT-5, GPT-5-mini, Gemini-2.5-Pro, Gemini-2.5-Flash). A stratified random sample of 250 model responses (50 per 0.2-score interval) was re-evaluated under identical protocols. Table~\ref{tab:interjudge} shows consistently high correlations, with Pearson ranging from 0.863 to 0.903 and Spearman from 0.866 to 0.901. Krippendorff's $\alpha = 0.867$ exceeds the conventional 0.80 threshold, indicating substantial agreement across models with different architectures.

Furthermore, to assess alignment with human judgments, the same 250-sample set was evaluated by four independent human annotators, who rated the appropriateness of GPT-5-Mini judgments on a 5-point scale. Agreement was high (Pearson $r = 0.820$, Spearman $\rho = 0.810$, $p < 0.001$), demonstrating that our judge provides a stable and human-aligned evaluation signal. Detailed analyses of low-agreement 
cases suggest that most discrepancies stem from superficial keyword matching or excessive leniency (examples in Appendix~\ref{app:judge_failures}).

\section{Related Work}

\paragraph{Evaluating VLMs.} As VLMs become more general-purposed, evaluation has shifted toward diagnostic suites that aim to separate recognition, OCR, and knowledge from higher-level reasoning and instruction following~\citep{liu2024mmbench,li2024seed, yu2024mmvet}. To better probe reasoning, several benchmarks target domain knowledge grounded with visual inputs~\citep{yue2024mmmu, yue2025mmmu, lu2023mathvista}.
This was rapidly followed by the Korean community, first by text benchmarks that measure Korean knowledge~\citep{son2023hae, son2025kmmlu, hong2025kmmlu}, then by multimodal benchmarks: KRETA, KViscuit, and KOFFVQA~\citep{hwang2025kreta,k-viscuit, KOFFVQA}. In addition, localized evaluation tools such as KMMB, KSEED, and KDTCBench have been released alongside Korean VLM development efforts~\citep{VARCO-VISION}. However, these benchmarks have already been saturated by older-generation models such as GPT-4o (e.g., KRETA~\citep{hwang2025kreta}: 84.6; K-VISCUIT~\citep{k-viscuit}: 89.5; K-MMB: 81.01; K-SEED: 76.98; K-DTCBench: 85.80~\citep{VARCO-VISION}), motivating the creation of a more challenging benchmark.

\paragraph{Query Underspecification.}
Underspecified or ambiguous queries are pervasive in conversational settings~\citep{rahmani2023survey}, forcing systems to choose between answering, hedging, or asking for missing constraints. Prior efforts to evaluate LLMs in ambiguity handling include AmbigQA~\citep{min2020ambigqa}, and clarification-focused resources such as ClariQ~\citep{aliannejadi2021building} and the ConvAI3 shared task~\citep{aliannejadi2020convai3}, which measure how effectively a system reduces uncertainty through clarification. More recently, QuestBench tests minimal question asking as information acquisition for underspecified reasoning~\citep{li2025questbench}. In the multimodal setting, ClearVQA evaluates whether models can ask image grounded clarification questions to resolve ambiguous visual queries~\citep{jian2025clearvqa}. Overall, however, multimodal resources for query underspecification remain scarce. To bridge this gap, we introduce \dataset, which further targets a niche and underexplored setting by focusing on underspecification in Korean language interactions with culturally grounded content and assumptions.

\section{Conclusion}

We introduce \dataset{}, a benchmark of 653 authentic Korean questions from real-life users, each paired with explicit rewrites. Our experiments show that query underspecification accounts for an 8--22 point drop in VLM performance. Retrieval-augmented prompting does not close this gap: search-augmented underspecified queries still underperform explicitated queries without search. We further find that many remaining failures reflect missing cultural knowledge rather than surface-level ambiguity. An English pilot study confirms that the explicitation effect generalizes cross-lingually, with smaller deltas attributable to fewer cultural knowledge barriers. Together, these findings highlight challenges that sanitized, clean-query benchmarks fail to capture.

\subsection*{Limitations}

Guided by a quality over quantity principle, our filtering procedure yields a 0.76\% survival rate. 
This aggressive filtering may exclude some informative edge cases; however, it should be noted that our goal is not to provide a comprehensive evaluation of Korean knowledge. Rather, we aim to study how LLM behavior changes under different levels of information density in user prompts. 
Furthermore, our web search augmentation analysis is also limited in scope, as it evaluates only OpenAI's web search, and results may differ with more advanced retrieval systems. 
However, based on our observations, the primary bottleneck appears to be less about the search API itself and more about the model's ability to extract and formulate meaningful questions grounded in the image and accompanying text. 
Our error annotation relies on an LLM judge, which may introduce systematic biases despite the high inter-judge agreement we observe.
Finally, while our English pilot (Section~\ref{sec:cross_lingual}) confirms that the explicitation effect generalizes beyond Korean, it is limited to 100 samples from a single platform ecosystem; a broader multilingual investigation across diverse languages and cultural contexts remains for future work.

\subsection*{Ethics and Data Governance}

This study received ethical approval from the Institutional Review Board of Hankuk University of Foreign Studies (HUFS-2510-015). All data were collected from publicly available Korean community platforms. We implemented a rigorous filtering process to exclude sensitive content, and all PII has been systematically removed. AI assistants (Claude and Gemini) were used for grammar editing and code debugging.

\subsubsection*{Acknowledgments}
This work was supported by the Hankuk University of Foreign Studies Research Fund (2025) and the TIPS Program (No. RS-2024-00512659) funded by the Ministry of SMEs and Startups (MSS), Korea.

\bibliography{custom}

\appendix
\onecolumn

\section*{Appendices}

\startcontents[sections]
\printcontents[sections]{l}{1}{\setcounter{tocdepth}{2}}

\clearpage

\section{Dataset Construction Details}

\subsection{Detailed Platform Descriptions}
\label{app:platforms}

We collected data from nine Korean online platforms representing diverse user communities and domain expertise. Table~\ref{tab:platforms} provides detailed information about each platform.

\begin{table*}[ht]
\centering
\small
\renewcommand{\arraystretch}{1.05} 
\resizebox{\textwidth}{!}{%
\begin{tabular}{@{}p{2.5cm}p{2.9cm}p{8.8cm}@{}}
\toprule
\textbf{Platform} & \textbf{Category} & \textbf{Description} \\
\midrule
Naver KnowledgeIn & General Q\&A & Korea's largest general Q\&A platform covering everyday queries, academic subjects, and technical issues \\
BRIC & Science Community & Specialized community for biological research and biotechnology with scientific discussions and professional knowledge sharing \\
Ruliweb & Gaming Community & Major gaming community covering video games, hardware reviews, game mechanics, and technical gaming issues \\
MonsterZym & Fitness Community & Fitness and bodybuilding community discussing workout routines, nutrition, supplements, and exercise techniques \\
Quasarzone & Hardware Community & Hardware enthusiast community focused on computer components, electronics, PC building, and technology reviews \\
i-Boss & Business Platform & Business and entrepreneurship platform for startup strategies, operations, marketing, and professional development \\
Inflearn & Coding Education & Online learning platform with community features for programming questions and coding experiences \\
Codeit & Coding Education & Coding education platform with forums for programming discussions and technical support \\
Okky & Developer Community & Developer community platform for programming discussions, career advice, and technical problem-solving \\
\bottomrule
\end{tabular}%
}
\caption{Korean online platforms used for data collection}
\label{tab:platforms}
\end{table*}

These platforms were selected to ensure comprehensive coverage of different user demographics, expertise levels, and domain-specific knowledge, reflecting the diversity of real-world multimodal questions Korean users encounter online.

\subsection{Platform-wise Filtering Statistics}
\label{app:platform_stats}

Table~\ref{tab:platform_results} provides a detailed breakdown of data collection and filtering across all platforms.

\begin{table*}[ht]
\centering
\small
\begin{tabular}{@{}lrrrrrrr@{}}
\toprule
Platform & Raw Data & Appropri. & Difficulty & Image Dep. & Human Val. & Final & Survival \\
\midrule
KnowledgeIn & 31,484 & 10,495 & 1,404 & 648 & 441 & 441 & 1.4\% \\
BRIC & 291 & 291 & 163 & 60 & 42 & 42 & 14.4\% \\
Ruliweb & 305 & 240 & 54 & 42 & 32 & 32 & 10.5\% \\
Coding & 27,896 & 8,369 & 837 & 198 & 135 & 135 & 0.5\% \\
MonsterZym & 3,090 & 3,090 & 2,234 & 8 & 6 & 6 & 0.2\% \\
Quasarzone & 2,986 & 896 & 90 & 22 & 15 & 15 & 0.5\% \\
i-Boss & 20,000 & 20,000 & 578 & 62 & 42 & 42 & 0.2\% \\
\midrule
Total & 86,052 & 43,381 & 5,360 & 1,040 & 713 & 653 & 0.76\% \\
\bottomrule
\end{tabular}
\caption{Detailed data collection and filtering statistics by platform (Stages 1–6). Coding platforms include Inflearn, Codeit, and Okky combined.}
\label{tab:platforms}
\label{tab:platform_results}
\end{table*}

\subsection{English Pilot Study Details}
\label{app:english_pilot}

To validate the cross-lingual generalizability of our findings, we construct an English pilot dataset by applying the same six-stage pipeline described in Section~\ref{subsec:pipeline}.

\paragraph{Data Source.}
We collect 2,954 image-containing Q\&A pairs from 12 Stack Exchange communities: Stack Overflow, Super User, Arqade, DIY Home Improvement, Biology, Gardening \& Landscaping, Motor Vehicle Maintenance, Cooking, Bicycles, Chemistry, Board \& Card Games, and Mathematics. All data are publicly available under CC BY-SA 4.0.

\paragraph{Filtering and Explicitation.}
All pipeline stages are applied without modification. Table~\ref{tab:english_pipeline} summarizes the filtering process. Image dependency verification is the most aggressive filter, removing over 60\% of candidates where images served as supplementary illustration rather than essential context. Each surviving question is explicitated following the same protocol as the Korean dataset (Appendix~\ref{app:explicitation_prompt}), with subsequent human verification. We randomly select 100 samples stratified by domain for the pilot evaluation.

\begin{table}[h]
\centering
\renewcommand{\arraystretch}{0.95}
\small
\begin{tabular}{lrc}
\toprule
\textbf{Stage} & \textbf{Remaining} & \textbf{Removed} \\
\midrule
Raw collection & 2,954 & -- \\
Image validation & 2,876 & $-$78 \\
Under-specification filter & 2,457 & $-$419 \\
Difficulty calibration & 2,326 & $-$131 \\
Image dependency verification & 887 & $-$1,439 \\
Checklist + explicitation & 887 & -- \\
Quality filtering & 627 & $-$260 \\
Stratified sampling & 100 & -- \\
\bottomrule
\end{tabular}
\caption{\footnotesize \textbf{English pilot filtering pipeline.}}
\label{tab:english_pipeline}
\end{table}

\paragraph{Dataset Statistics.}
The final 100 samples contain 168 images (avg 1.7 per question) and an average of 3.4 checklist items. Table~\ref{tab:english_pilot_dist} shows the domain distribution.

\begin{table}[h]
\centering
\renewcommand{\arraystretch}{0.95}
\small
\begin{tabular}{lcc}
\toprule
\textbf{Domain} & \textbf{\# Items} & \textbf{\%} \\
\midrule
Natural Objects / Science & 23 & 23.0 \\
Transportation & 18 & 18.0 \\
Gaming / Entertainment & 13 & 13.0 \\
Daily Life / Machinery & 12 & 12.0 \\
Coding & 11 & 11.0 \\
IT / Computer & 9 & 9.0 \\
Daily Life & 7 & 7.0 \\
Science & 4 & 4.0 \\
Mathematics & 3 & 3.0 \\
\midrule
\textbf{Total} & \textbf{100} & \textbf{100.0} \\
\bottomrule
\end{tabular}
\caption{\footnotesize \textbf{Domain distribution of the English pilot dataset.}}
\label{tab:english_pilot_dist}
\end{table}

\paragraph{Per-Domain Results.}
Table~\ref{tab:english_pilot_domain} presents per-domain explicitation effects averaged across all four models. Consistent with the Korean results, Coding shows the largest gain (+14.4), while visually grounded domains such as Natural Objects show smaller improvements.

\begin{table}[h]
\centering
\renewcommand{\arraystretch}{0.95}
\small
\begin{tabular}{lcccc}
\toprule
\textbf{Domain} & \textbf{$n$} & \textbf{Orig} & \textbf{Expl} & \textbf{$\Delta$} \\
\midrule
Coding & 11 & 57.5 & 71.8 & +14.4 \\
Gaming / Ent. & 13 & 40.0 & 49.0 & +9.0 \\
Daily Life & 7 & 47.3 & 55.6 & +8.3 \\
Science & 4 & 49.8 & 57.8 & +8.0 \\
Transportation & 18 & 51.1 & 56.5 & +5.4 \\
Nat.\ Objects / Sci. & 23 & 55.4 & 58.3 & +2.9 \\
IT / Computer & 9 & 55.4 & 58.1 & +2.7 \\
Mathematics & 3 & 49.4 & 50.6 & +1.1 \\
Daily Life / Mach. & 12 & 60.1 & 56.8 & $-$3.3 \\
\bottomrule
\end{tabular}
\caption{\footnotesize \textbf{English pilot: per-domain explicitation effects.} Scores averaged across all four evaluated models.}
\label{tab:english_pilot_domain}
\end{table}

\section{Pipeline Prompts}

\subsection{Stage~2 (Safety, Objectivity, Temporal)}
\label{app:stage2_prompts}

We used three LLM-based filters in Stage~2: content safety, objectivity, and temporal dependency. 
Below we excerpt only the core exclusion criteria from the prompts (full wording omitted).

\subsubsection{Content Safety}

\begin{tcolorbox}[fontupper=\small]
Mark as inappropriate if the question–image pair includes:
\begin{itemize}
\item Political content (politicians, parties, elections, political opinions)
\item Religious advocacy/criticism or conflicts
\item Hate/discrimination
\item Suicide or self-harm; sensitive mental-health topics
\item Sexual/adult content, nudity, explicit innuendo
\end{itemize}
\end{tcolorbox}

\subsubsection{Objectivity}

\begin{tcolorbox}[fontupper=\small]
Mark as inappropriate if the pair is subjective or ambiguous, e.g.:
\begin{itemize}
\item Preference/aesthetic judgments (“pretty/ugly”, “which outfit is nicer?”)
\item Suitability/personal advice without criteria
\item Moral/intentionality speculation (“who is wrong?”, “good person?”)
\item Multiple valid interpretations or unverifiable answers
\end{itemize}
\end{tcolorbox}

\subsubsection{Temporal Dependency}

\begin{tcolorbox}[fontupper=\small]
Mark as inappropriate if the pair requires time-specific information, e.g.:
\begin{itemize}
\item “today/now” weather, traffic, store hours, last train
\item Current events or status queries (“is it open now?”, “stock price today?”)
\item Questions that become invalid/meaningless as time passes
\end{itemize}
\end{tcolorbox}

\subsection{Stage~4 Prompt Excerpt (Image Dependency Rubric)}\label{app:stage4_prompt}

\begin{tcolorbox}[fontupper=\small]
\textbf{Input:} (Q), model answer with image, model answer without image, optional gold answer snippet.\\
\textbf{Task:} Compare the two answers and decide image dependency.

\textbf{Decision labels}
\begin{itemize}
  \item \textsc{IMAGE\_REQUIRED}: with-image answer is substantially more accurate/specific; text-only answer is vague, incorrect, or explicitly requests the image.
  \item \textsc{NO\_IMAGE\_NEEDED}: both answers are comparable in correctness and specificity without relying on visual cues.
  \item \textsc{UNCERTAIN}: evidence is inconclusive (e.g., partial improvements or conflicting signals).
\end{itemize}

\textbf{Scoring (1--5 quality gap)}
\begin{itemize}
  \item 1: negligible difference; 3: clear but moderate gain; 5: decisive gain (critical visual details).
\end{itemize}

\textbf{Output (natural language)}
\begin{itemize}
  \item \texttt{Judgment:} \textsc{IMAGE\_REQUIRED} / \textsc{NO\_IMAGE\_NEEDED} / \textsc{UNCERTAIN}
  \item \texttt{Reason:} brief comparison citing concrete differences
  \item \texttt{QualityGap:} integer in \{1,2,3,4,5\}
\end{itemize}
\end{tcolorbox}

\subsection{Stage~5 (Checklist Generation)}
\label{app:stage5_prompt}

This appendix provides the instruction prompt used for checklist generation 
along with illustrative examples of the resulting decompositions. 
We used GPT-4-mini to derive structured criteria directly from reference answers that users found satisfactory. 
These checklists therefore represent strict, human-aligned evaluation standards: 
a model must satisfy all listed criteria to be considered correct.

\newcommand{\SimpleCard}[2]{%
  \fcolorbox{black!15}{gray!4}{%
    \parbox[t][0.10\textheight][t]{\linewidth}{%
      #1\par\vspace{-0.05em} 
      {\setlist[itemize]{leftmargin=1.4em, itemsep=0pt, topsep=0pt}
       #2}%
    }%
  }%
}
\begin{figure}[t]
\centering
\footnotesize
\begin{minipage}[t]{0.49\linewidth}
  \SimpleCard{Game (Stardew Valley) \\ \emph{“What is the circled item in the screenshot?”}}{%
    \begin{itemize}
      \item Identify circled item as a sap tap (\begin{CJK*}{UTF8}{mj}수액 채취기\end{CJK*})
      \item Mention install only on fully grown trees
      \item Explain how to obtain/craft it
      \item Note sap can be collected after time
    \end{itemize}
  }
\end{minipage}\hspace{10pt}
\begin{minipage}[t]{0.46\linewidth}
  \SimpleCard{Economics/Management \\ \emph{“Cost allocation: is S2 missing 100{,}000?”}}{%
    \begin{itemize}
      \item Provide correct S1/S2 values
      \item Reset self-allocation entries to zero
      \item Derive allocation ratios (0.5F, 0.4M)
    \end{itemize}
  }
\end{minipage}

\vspace{5pt}

\begin{minipage}[t]{0.49\linewidth}
  \SimpleCard{Daily Life \\ \emph{“Is this ceiling tile asbestos?”}}{%
    \begin{itemize}
      \item Identify material as gypsum, not asbestos
      \item Explain gypsum board contains no asbestos
      \item Explicitly name ``\begin{CJK*}{UTF8}{mj}석고텍스\end{CJK*}''
      \item Assure user it is safe
    \end{itemize}
  }
\end{minipage}\hspace{10pt}
\begin{minipage}[t]{0.46\linewidth}
  \SimpleCard{Science \\ \emph{“Why does neutron mass ratio decrease?”}}{%
    \begin{itemize}
      \item Explain neutron beta decay
      \item Clarify neutrons inside He nucleus
      \item Relate $x$-axis to cosmic cooling
      \item Interpret H:He ratio $\approx 3{:}1$
    \end{itemize}
  }
\end{minipage}
\caption{Examples of checklist decomposition across domains, generated in Stage~5. For brevity, the checklists shown here are abbreviated; full checklists typically contain 1--5 criteria per item.}
\label{fig:checklist_cards}
\end{figure}

\newpage
\subsection{Query Explicitation Prompt}
\label{app:explicitation_prompt}

The following prompt was used with GPT-5.1 (web search enabled) to generate explicitated versions of under-specified queries.

\begin{figure}[H]
\begin{tcolorbox}[colback=gray!5, colframe=gray!50, boxrule=0.5pt, arc=2pt]
\small
You rewrite incomplete, ambiguous, or context-dependent questions into clear, fully self-contained questions. Your goal is to produce a rewritten question that can be understood and answered on its own, without requiring prior conversation or hidden context. Preserve the original intent, scope, and tone of the question. Do NOT answer the question.

\textbf{Rules:}

\textbf{1. Intent and scope preservation}
\begin{itemize}[leftmargin=1.5em, itemsep=0pt, topsep=2pt]
\item Preserve what the original question is asking and its level of specificity.
\item Do not broaden or narrow the scope.
\item Do not generalize away concrete entities or situations implied by the original question or answer.
\end{itemize}

\textbf{2. Essential context inclusion}
\begin{itemize}[leftmargin=1.5em, itemsep=0pt, topsep=2pt]
\item Explicitly include essential context if it is implied or required to understand the question, such as: the relevant domain or subject; the specific scenario, task, or situation involved; named entities (e.g., people, organizations, characters, locations); concrete objects, items, or targets referenced.
\item Avoid vague references such as ``this,'' ``that,'' ``here,'' ``the scene,'' or ``the above.''
\end{itemize}

\textbf{3. Search usage}
\begin{itemize}[leftmargin=1.5em, itemsep=0pt, topsep=2pt]
\item You may use search ONLY to identify widely accepted proper nouns (e.g., titles, names, commonly used labels) that are strongly implied by the original question or associated answer.
\item Do NOT use search to introduce new mechanics, steps, conditions, quantities, or interpretations.
\item Do NOT resolve ambiguity by inventing details.
\end{itemize}

\textbf{4. Handling missing or ambiguous information}
\begin{itemize}[leftmargin=1.5em, itemsep=0pt, topsep=2pt]
\item If critical context cannot be inferred with high confidence, include a brief clarifying placeholder inside the question, such as: [SPECIFY: missing detail].
\item Do not attempt to guess or ``fix'' the question beyond what the inputs support.
\end{itemize}

\textbf{5. Image usage (if an image is provided)}
\begin{itemize}[leftmargin=1.5em, itemsep=0pt, topsep=2pt]
\item You may incorporate information visible in the image to clarify the question.
\item The rewritten question must remain answerable without viewing the image.
\item Do not exhaustively describe the image or convert all visual details into text.
\item Include only visual information that is essential to understanding the question.
\end{itemize}

\textbf{6. Language and style}
\begin{itemize}[leftmargin=1.5em, itemsep=0pt, topsep=2pt]
\item Maintain a tone consistent with the original question.
\item Do not unnecessarily formalize or casualize the language.
\item Remove slang, conversational fillers, and vague references that reduce clarity.
\end{itemize}

\textbf{Output requirements:} Output ONLY the rewritten question text. No explanations, no bullet points, no headers. Do not include meta-instructions or commentary.
\end{tcolorbox}
\end{figure}

\section{Human Annotation}
\subsection{Annotation Guidelines}
\label{app:guidelines}

Seven Korean-speaking annotators conducted human validation in three phases using custom web-based tools.

\subsubsection{Phase 1: Conservative Filtering}

\begin{figure}[H]
\centering
\includegraphics[width=0.82\linewidth]{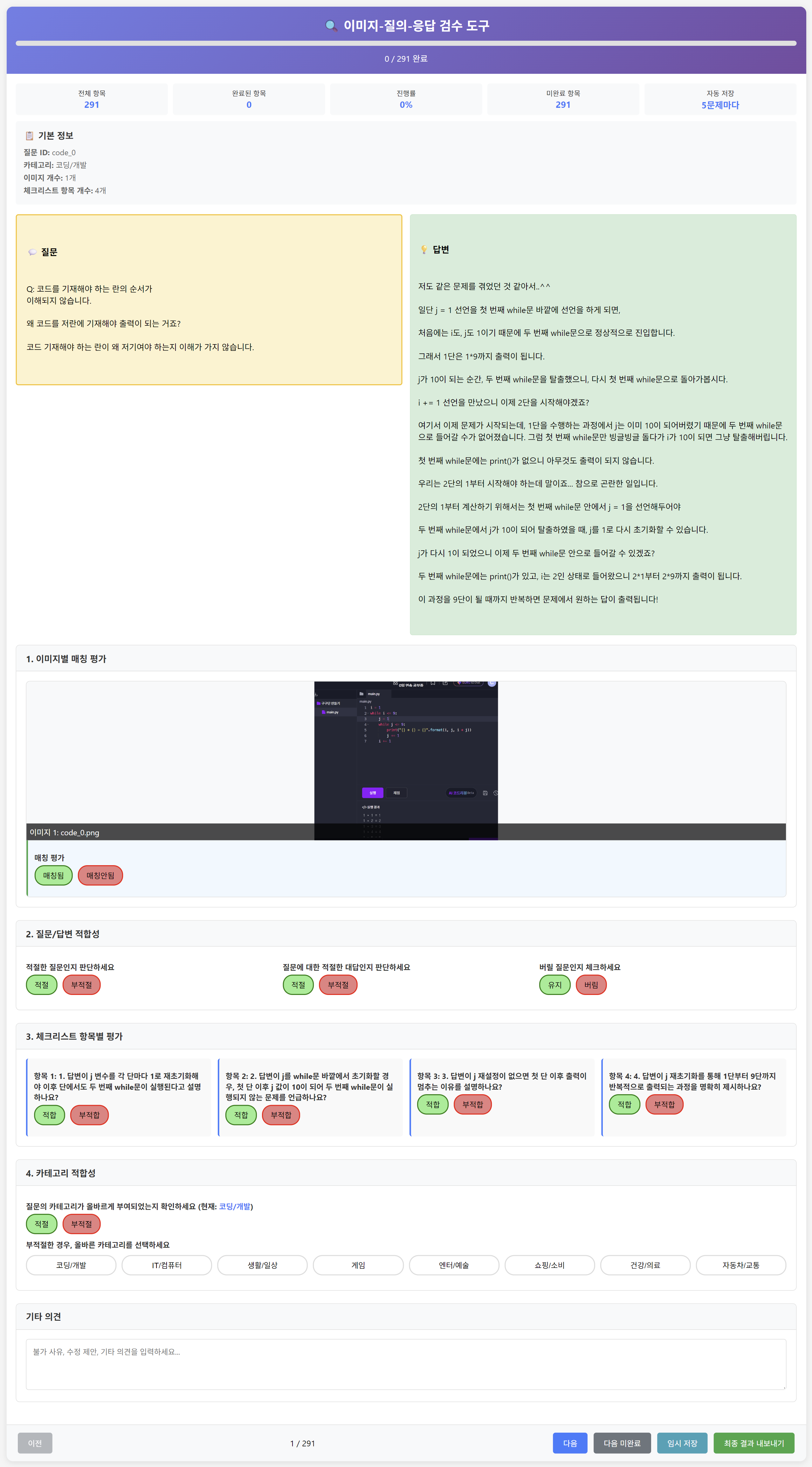} 
\caption{Screenshot of our Phase 1 annotation tool. The interface (shown in Korean) allowed annotators to assess image relevance, question/answer appropriateness, checklist accuracy, and category assignment.} 
\label{fig:annotation_tool}
\end{figure}

Using the annotation interface shown in Figure~\ref{fig:annotation_tool}, annotators independently reviewed each item along five dimensions, removing any item flagged by at least one annotator:

\begin{itemize}
\item \textbf{Image-Question Relevance}: Assess whether images provide essential visual information required to answer the question.
\item \textbf{Question-Answer Quality}: Evaluate question clarity, answerability, and reference answer accuracy.
\item \textbf{Checklist Validation}: Review each LLM-generated checklist item for necessity, clarity, and completeness.
\item \textbf{Category Appropriateness}: Verify correct classification into one of 13 domain categories.
\item \textbf{Overall Assessment}: Flag items with fundamental issues such as inappropriate content or unsolvable questions.
\end{itemize}

\subsubsection{Phase 2: Refinement}

Three annotators refined surviving items through a separate annotation interface:

\begin{itemize}
\item \textbf{Question Rewriting}: Rewrite unclear or ambiguous questions while preserving original intent and scope.
\item \textbf{Checklist Revision}: Evaluate each LLM-generated checklist item for appropriateness, revising unclear criteria or removing items not grounded in the original question–image pair.
\item \textbf{Category Re-assignment}: Re-assign categories where the original classification was incorrect, with option to propose new categories.
\end{itemize}

\subsubsection{Phase 3: Final Audit}

One senior annotator consolidated categories across the dataset and verified cross-item consistency.

\newpage

\subsection{LLM Judge Failure Cases}
\label{app:judge_failures}

Table~\ref{tab:judge_failure_examples} presents representative examples of human annotator feedback for inappropriate judge evaluations, revealing systematic failure patterns.

\begin{table}[ht]
\centering
\small
\begin{tabular}{cp{11cm}}
\toprule
\textbf{Rating} & \textbf{Human Reasoning (translated)} \\
\midrule
Very Inappropriate & "Judge awarded points based on superficial word matching rather than actual checklist compliance" \\
\midrule
Inappropriate & "Judge gave 1 point despite response not addressing checklist criteria, incorrectly interpreting explicit mention as meeting requirements" \\
\midrule
Inappropriate & "Checklists 1,2,4 satisfied. Item 3 not clearly inappropriate but ambiguous and open to interpretation" \\
\midrule
Inappropriate & "Even if intent aligns with checklist, response lacks clarity and remains ambiguous" \\
\midrule
Inappropriate & "Judge overlooked insufficient explanations that clearly failed checklist requirements" \\
\bottomrule
\end{tabular}
\caption{Representative human feedback explaining inappropriate judge ratings.}
\label{tab:judge_failure_examples}
\end{table}

Analysis reveals judge failures primarily stem from: (1) superficial keyword matching without semantic understanding, (2) excessive leniency toward incomplete responses, and (3) difficulty distinguishing between implicit intent and explicit satisfaction of requirements.

\section{LLM-as-Judge Prompt}
\label{app:judge_prompt}

This appendix provides the full prompt used for the checklist-based evaluation by the GPT-5-Mini judge. The prompt enforces explicitness, evidence grounding, and consistent scoring across items. For reproducibility, we include the full decision rules, evidence policy, and output format constraints.

\begin{tcolorbox}[breakable, enhanced, colback=gray!3, colframe=black!15, fontupper=\small]

\vspace{0.3em}
[GOAL]
Given a Question, Response, and a natural-language Checklist,
decide for each checklist item whether the Response explicitly satisfies it:
met = 1, partially met = 0.5, not met = 0.
Final score = (\# met) / (total checklist items).

\vspace{0.6em}
[INPUT]

\vspace{0.6em}
[Question]
{{QUESTION}}

\vspace{0.6em}
[Response]
{{RESPONSE}}

\vspace{0.6em}
[Checklist]
{{CHECKLIST}}
Treat each string as one criterion. Remove numbering such as "1." or "2)".

\vspace{0.6em}
[DECISION RULES]

\begin{itemize}
  \item \textbf{Use only the Response text.}  
        No outside knowledge or assumptions. If uncertain → 0.

  \item \textbf{Explicitness:}  
        direct fulfillment = 1, implicit or suggestive = 0.5, otherwise = 0.

  \item \textbf{Completeness ("all / every / complete"):}  
        explicit = 1, implied = 0.5, absent = 0.

  \item \textbf{Method requirements:}  
        actionable steps = 1, vague = 0.5, absent = 0.

  \item \textbf{"Various / multiple types":}  
        $\ge 2$ specific types = 1, vague or 1 type = 0.5, none = 0.

  \item \textbf{Synonyms:}  
        unambiguous = 1, ambiguous = 0.5, different meaning = 0.
\end{itemize}

\vspace{0.6em}
[EVIDENCE POLICY]

\begin{itemize}
  \item For \textbf{1} or \textbf{0.5}: include a 10–60 character direct quote.
  \item For \textbf{0}: provide a brief explanation.
  \item Each item must include: \emph{evidence → explanation → met}.
\end{itemize}

\vspace{0.6em}
[OUTPUT FORMAT — STRICT]

\hspace*{2em}<evidence>\\
\hspace*{2em}Item 1:\\
\hspace*{4em}evidence: "…direct quote from Response…" \\
\hspace*{4em}explanation: Brief justification referencing criteria\\
\hspace*{4em}met: 0 | 0.5 | 1
\\[0.5em]

\hspace*{2em}Item 2:\\
\hspace*{4em}evidence: "…"\\
\hspace*{4em}explanation: …\\
\hspace*{4em}met: 0 | 0.5 | 1\\

\hspace*{2em}... (repeat)\\
\hspace*{2em}</evidence>
\\[0.5em]

\hspace*{2em}<score>\\
\hspace*{4em}K/N\\
\hspace*{2em}</score>

\vspace{0.6em}
[NOTES]
\begin{itemize}
\item Output only the two blocks above.
\item No code fences or additional prose.
\end{itemize}

\end{tcolorbox}

\label{app:full_results}
\begin{landscape}
\begin{table*}[p]
\centering
\scriptsize
\setlength{\tabcolsep}{4pt}
\renewcommand{\arraystretch}{0.8}
\begin{tabular}{lcccccccccccccc}
\toprule
Model & IT & Health & Game & Econ & Sci & Mach & Daily & Shop & Math & Ent & Trans & Nature & Code & Avg \\
\midrule
\multicolumn{15}{l}{\textit{Proprietary Models}} \\
Gemini 2.5 Pro & $50.73_{1.63}$ & $62.17_{2.33}$ & $36.67_{1.75}$ & $51.09_{1.72}$ & $39.93_{1.43}$ & $56.22_{4.32}$ & $51.32_{2.57}$ & $46.00_{5.15}$ & $60.94_{1.36}$ & $44.37_{1.18}$ & $57.69_{2.81}$ & $53.45_{0.57}$ & $50.91_{0.92}$ & $48.54_{0.18}$ \\

Gemini 2.5 Flash & $42.98_{0.34}$ & $56.10_{3.95}$ & $26.70_{1.04}$ & $48.05_{6.64}$ & $39.62_{3.14}$ & $45.86_{1.70}$ & $44.59_{2.10}$ & $46.13_{5.19}$ & $51.14_{3.87}$ & $31.92_{3.63}$ & $48.12_{2.37}$ & $44.37_{0.99}$ & $39.26_{2.25}$ & $41.05_{1.38}$ \\
Gemini 2.5 Flash Lite & $25.92_{1.82}$ & $43.54_{4.48}$ & $17.97_{1.92}$ & $38.84_{3.82}$ & $41.73_{1.47}$ & $38.30_{3.10}$ & $30.67_{0.88}$ & $28.82_{2.38}$ & $45.62_{7.49}$ & $18.82_{0.68}$ & $34.10_{3.78}$ & $27.16_{0.32}$ & $32.63_{2.66}$ & $30.29_{0.42}$ \\

GPT 5 & $59.95_{2.01}$ & $62.61_{2.59}$ & $32.34_{2.08}$ & $58.41_{1.63}$ & $36.31_{1.60}$ & $52.85_{4.72}$ & $46.93_{2.83}$ & $55.96_{3.14}$ & $54.70_{4.54}$ & $33.80_{2.20}$ & $54.97_{2.43}$ & $53.42_{1.23}$ & $55.07_{1.24}$ & $48.01_{0.32}$ \\
GPT 5 Mini & $49.59_{3.74}$ & $60.45_{2.40}$ & $29.22_{1.71}$ & $50.19_{5.53}$ & $52.49_{0.44}$ & $51.68_{1.47}$ & $50.28_{4.75}$ & $44.33_{4.96}$ & $58.19_{3.94}$ & $25.54_{2.20}$ & $49.22_{3.11}$ & $41.17_{2.73}$ & $57.02_{0.53}$ & $45.21_{1.21}$ \\
GPT 5 Nano & $22.99_{2.64}$ & $45.98_{1.02}$ & $10.46_{1.71}$ & $24.81_{0.65}$ & $11.47_{1.21}$ & $26.59_{7.45}$ & $21.49_{1.41}$ & $26.42_{3.26}$ & $23.56_{6.12}$ & $12.81_{0.67}$ & $26.17_{1.63}$ & $25.27_{1.60}$ & $32.84_{4.80}$ & $21.22_{0.46}$ \\

Grok 4 & $39.64_{1.89}$ & $36.96_{1.16}$ & $29.00_{1.49}$ & $44.44_{2.79}$ & $40.70_{1.13}$ & $47.63_{1.86}$ & $40.57_{1.60}$ & $36.73_{1.65}$ & $22.09_{2.86}$ & $24.77_{1.78}$ & $50.43_{4.90}$ & $30.29_{1.28}$ & $39.02_{0.20}$ & $36.08_{0.53}$ \\
\midrule
\multicolumn{15}{l}{\textit{Open-source Models}} \\
\multicolumn{15}{l}{\quad\textit{Mistral/Pixtral Family}} \\
Mistral Medium 3.1 & $24.77_{2.76}$ & $37.01_{5.96}$ & $16.01_{1.49}$ & $28.48_{2.48}$ & $33.70_{1.23}$ & $34.14_{1.20}$ & $24.41_{1.06}$ & $25.22_{2.51}$ & $38.99_{3.41}$ & $11.46_{2.32}$ & $25.46_{2.65}$ & $19.62_{2.62}$ & $31.09_{2.35}$ & $24.86_{0.98}$ \\
Pixtral Large & $19.09_{1.82}$ & $35.09_{2.33}$ & $11.33_{1.33}$ & $24.32_{3.36}$ & $27.40_{2.29}$ & $23.41_{1.21}$ & $24.16_{1.09}$ & $19.01_{4.66}$ & $19.89_{1.10}$ & $11.54_{2.50}$ & $21.93_{1.34}$ & $18.08_{0.62}$ & $22.64_{0.67}$ & $20.10_{0.41}$ \\
Mistral Small 24B  & $15.38_{1.93}$ & $25.07_{4.25}$ & $7.00_{1.35}$ & $22.29_{1.84}$ & $20.47_{1.46}$ & $21.53_{2.01}$ & $13.07_{2.67}$ & $15.34_{3.68}$ & $18.57_{1.81}$ & $7.76_{1.09}$ & $13.84_{3.49}$ & $10.61_{1.77}$ & $16.36_{2.46}$ & $14.43_{0.41}$ \\
Pixtral 12B & $8.76_{0.77}$ & $24.19_{3.58}$ & $6.74_{0.94}$ & $17.49_{0.53}$ & $14.12_{0.11}$ & $16.46_{2.65}$ & $11.66_{2.40}$ & $11.44_{1.34}$ & $6.83_{2.27}$ & $6.17_{0.35}$ & $15.06_{2.39}$ & $9.60_{0.46}$ & $12.94_{2.80}$ & $11.20_{0.02}$ \\
\multicolumn{15}{l}{\quad\textit{Google Gemma Family}} \\
Gemma 3 27B & $20.31_{1.18}$ & $40.90_{1.49}$ & $13.75_{1.55}$ & $31.71_{3.21}$ & $34.93_{1.52}$ & $27.36_{6.28}$ & $26.72_{1.12}$ & $24.07_{2.01}$ & $23.85_{2.74}$ & $9.43_{1.30}$ & $20.66_{2.62}$ & $18.61_{0.40}$ & $20.81_{2.15}$ & $22.53_{0.28}$ \\
Gemma 3 12B & $15.15_{0.69}$ & $36.60_{1.32}$ & $10.52_{1.44}$ & $27.91_{1.30}$ & $28.79_{1.39}$ & $27.44_{3.60}$ & $19.20_{1.27}$ & $22.40_{1.47}$ & $17.25_{2.89}$ & $7.23_{1.12}$ & $21.01_{1.65}$ & $13.43_{1.47}$ & $23.41_{0.13}$ & $18.76_{0.63}$ \\
Gemma 3 4B & $12.43_{1.63}$ & $34.23_{1.08}$ & $8.91_{0.96}$ & $19.67_{4.37}$ & $22.50_{0.12}$ & $21.25_{1.33}$ & $15.59_{0.87}$ & $18.21_{1.21}$ & $13.54_{2.63}$ & $6.84_{1.10}$ & $19.56_{2.12}$ & $14.68_{1.08}$ & $13.45_{0.88}$ & $15.47_{0.78}$ \\
\multicolumn{15}{l}{\quad\textit{AIDC-AI Ovis2 Family}} \\
Ovis2-34B & $15.90_{1.35}$ & $40.15_{2.16}$ & $9.87_{0.77}$ & $19.44_{0.45}$ & $23.97_{0.56}$ & $29.46_{1.47}$ & $19.43_{0.58}$ & $20.27_{3.31}$ & $22.91_{2.46}$ & $9.18_{1.41}$ & $21.86_{2.89}$ & $18.77_{1.37}$ & $16.78_{0.26}$ & $18.50_{0.03}$ \\
Ovis2-16B & $11.20_{1.67}$ & $38.98_{0.75}$ & $8.08_{0.18}$ & $21.58_{1.27}$ & $24.68_{0.80}$ & $23.94_{3.50}$ & $21.20_{3.52}$ & $14.83_{3.00}$ & $24.32_{1.31}$ & $8.72_{1.57}$ & $20.21_{0.84}$ & $16.47_{0.63}$ & $16.12_{1.92}$ & $17.18_{0.50}$ \\
Ovis2-8B & $9.80_{1.30}$ & $33.62_{1.54}$ & $6.07_{0.30}$ & $19.18_{3.28}$ & $19.45_{1.85}$ & $21.02_{1.98}$ & $18.37_{1.83}$ & $13.51_{1.33}$ & $19.81_{5.29}$ & $8.04_{0.53}$ & $17.42_{3.08}$ & $13.17_{0.35}$ & $14.77_{1.80}$ & $14.46_{0.37}$ \\
Ovis2-4B & $6.76_{1.75}$ & $23.66_{3.93}$ & $6.00_{0.27}$ & $15.89_{2.76}$ & $16.16_{1.17}$ & $17.05_{3.15}$ & $16.43_{1.51}$ & $10.68_{2.89}$ & $13.16_{0.84}$ & $7.17_{0.50}$ & $17.65_{3.01}$ & $14.26_{0.58}$ & $8.31_{1.00}$ & $12.18_{0.11}$ \\
Ovis2-2B & $6.14_{0.22}$ & $16.10_{1.01}$ & $5.30_{0.83}$ & $13.74_{2.34}$ & $12.24_{1.70}$ & $13.64_{4.43}$ & $11.99_{1.14}$ & $11.27_{2.01}$ & $6.57_{1.32}$ & $7.28_{0.64}$ & $11.33_{2.19}$ & $9.73_{0.56}$ & $8.98_{3.88}$ & $9.54_{0.22}$ \\
Ovis2-1B & $4.83_{0.91}$ & $12.62_{2.58}$ & $4.74_{0.31}$ & $8.07_{1.07}$ & $7.52_{0.71}$ & $5.95_{1.12}$ & $8.03_{0.98}$ & $8.11_{1.97}$ & $6.57_{2.40}$ & $5.05_{0.98}$ & $8.10_{2.55}$ & $6.80_{1.38}$ & $4.43_{1.13}$ & $6.52_{0.25}$ \\
\multicolumn{15}{l}{\quad\textit{OpenGVLab InternVL3.5 Family}} \\
InternVL3.5 38B & $14.94_{0.63}$ & $30.95_{4.82}$ & $9.09_{1.57}$ & $24.85_{1.52}$ & $28.79_{0.27}$ & $20.90_{4.44}$ & $19.25_{1.90}$ & $18.40_{0.19}$ & $24.54_{2.47}$ & $8.53_{0.17}$ & $21.10_{0.84}$ & $16.41_{1.98}$ & $14.76_{2.12}$ & $18.01_{0.39}$ \\
InternVL3.5 14B & $15.50_{2.05}$ & $26.81_{4.46}$ & $8.26_{1.12}$ & $20.72_{0.96}$ & $24.64_{1.18}$ & $17.41_{3.95}$ & $14.67_{1.98}$ & $17.70_{3.09}$ & $26.45_{1.63}$ & $7.74_{0.53}$ & $15.76_{1.58}$ & $12.05_{1.07}$ & $19.72_{3.13}$ & $16.04_{0.37}$ \\
InternVL3.5 8B & $10.22_{1.08}$ & $23.11_{3.12}$ & $7.14_{0.57}$ & $20.44_{1.87}$ & $20.14_{1.89}$ & $16.16_{3.35}$ & $11.27_{1.56}$ & $11.99_{2.92}$ & $22.96_{2.08}$ & $5.29_{0.79}$ & $12.68_{1.73}$ & $12.57_{0.25}$ & $13.01_{1.22}$ & $13.16_{0.82}$ \\
InternVL3.5 4B & $7.70_{1.15}$ & $23.33_{0.30}$ & $7.72_{0.52}$ & $19.71_{1.60}$ & $23.20_{1.24}$ & $18.84_{0.42}$ & $15.11_{1.52}$ & $14.98_{2.03}$ & $25.72_{2.64}$ & $6.48_{0.96}$ & $13.83_{1.36}$ & $11.78_{1.37}$ & $14.90_{2.25}$ & $14.09_{0.28}$ \\
InternVL3.5 2B & $5.32_{0.25}$ & $20.86_{4.13}$ & $5.24_{0.34}$ & $15.50_{1.86}$ & $16.05_{0.87}$ & $12.69_{2.87}$ & $8.94_{1.54}$ & $7.69_{1.60}$ & $14.18_{1.71}$ & $5.63_{1.26}$ & $10.14_{3.14}$ & $7.03_{0.51}$ & $9.07_{2.28}$ & $9.48_{0.49}$ \\
InternVL3.5 1B & $3.21_{0.43}$ & $7.94_{2.99}$ & $3.39_{0.09}$ & $10.32_{0.07}$ & $9.12_{0.32}$ & $5.74_{0.58}$ & $3.29_{1.02}$ & $7.79_{1.30}$ & $10.22_{1.53}$ & $3.24_{0.57}$ & $7.31_{1.05}$ & $2.93_{0.74}$ & $5.64_{0.43}$ & $5.43_{0.13}$ \\
\midrule
\multicolumn{15}{l}{\quad\textit{Qwen2.5-VL Family}} \\
Qwen2.5 VL 72B & $16.53_{1.36}$ & $31.30_{1.38}$ & $11.80_{2.24}$ & $25.55_{1.06}$ & $28.46_{2.62}$ & $23.55_{1.42}$ & $19.72_{0.38}$ & $25.86_{3.14}$ & $32.32_{7.22}$ & $9.97_{0.45}$ & $21.02_{2.62}$ & $19.36_{0.79}$ & $25.31_{1.59}$ & $20.58_{0.80}$ \\
Qwen2.5 VL 7B & $10.33_{0.70}$ & $21.04_{4.51}$ & $5.95_{1.26}$ & $18.96_{1.05}$ & $20.49_{3.89}$ & $18.50_{3.79}$ & $13.70_{0.92}$ & $17.00_{4.00}$ & $13.26_{4.07}$ & $6.71_{0.28}$ & $14.06_{1.66}$ & $12.35_{0.74}$ & $13.28_{2.86}$ & $13.15_{0.86}$ \\
Qwen2.5 VL 3B & $6.08_{2.15}$ & $18.49_{3.90}$ & $2.82_{0.44}$ & $12.76_{1.17}$ & $11.54_{1.70}$ & $13.76_{2.51}$ & $9.22_{0.16}$ & $6.89_{1.47}$ & $10.14_{0.98}$ & $4.88_{0.18}$ & $10.31_{3.46}$ & $7.85_{0.38}$ & $6.54_{0.84}$ & $8.20_{0.36}$ \\
\multicolumn{15}{l}{\quad\textit{Qwen3-VL Family}} \\
Qwen3-VL-235B-A22B-Instruct & $37.75_{2.29}$ & $54.44_{3.96}$ & $23.28_{1.93}$ & $43.16_{3.45}$ & $51.51_{1.76}$ & $47.42_{4.00}$ & $39.14_{2.65}$ & $40.98_{4.03}$ & $54.31_{4.08}$ & $22.75_{2.42}$ & $36.33_{2.92}$ & $37.44_{1.74}$ & $40.10_{3.23}$ & $38.41_{0.76}$ \\
Qwen3-VL-235B-A22B-Thinking & $34.19_{2.38}$ & $52.12_{4.01}$ & $23.97_{1.87}$ & $47.30_{3.12}$ & $49.19_{1.91}$ & $38.37_{3.97}$ & $30.02_{2.49}$ & $34.18_{3.68}$ & $56.51_{3.95}$ & $20.29_{2.32}$ & $34.12_{2.80}$ & $33.04_{1.70}$ & $34.87_{3.16}$ & $35.47_{0.75}$ \\
Qwen3-VL-32B-Instruct & $36.74_{2.17}$ & $56.30_{3.78}$ & $18.13_{1.62}$ & $41.29_{3.18}$ & $51.39_{1.86}$ & $41.73_{3.76}$ & $34.28_{2.47}$ & $43.38_{3.72}$ & $60.92_{3.77}$ & $19.67_{1.87}$ & $36.02_{3.15}$ & $34.43_{1.66}$ & $32.25_{3.06}$ & $36.08_{0.73}$ \\
Qwen3-VL-32B-Thinking & $33.92_{2.36}$ & $52.39_{4.28}$ & $19.76_{1.72}$ & $38.21_{2.96}$ & $51.94_{1.75}$ & $35.66_{3.53}$ & $29.38_{2.40}$ & $38.16_{3.86}$ & $64.57_{3.65}$ & $19.19_{2.15}$ & $35.57_{3.14}$ & $35.04_{1.72}$ & $37.59_{3.04}$ & $35.49_{0.74}$ \\
Qwen3-VL-30B-A3B-Thinking & $36.19_{2.23}$ & $56.38_{3.50}$ & $18.06_{1.73}$ & $38.44_{3.21}$ & $49.92_{1.87}$ & $38.48_{3.81}$ & $32.34_{2.65}$ & $37.99_{3.87}$ & $68.69_{3.71}$ & $17.81_{2.13}$ & $37.40_{2.63}$ & $35.46_{1.66}$ & $29.95_{3.04}$ & $35.41_{0.74}$ \\
Qwen3-VL-30B-A3B-Instruct & $31.13_{2.26}$ & $54.71_{3.46}$ & $18.86_{1.65}$ & $42.02_{3.29}$ & $40.40_{1.66}$ & $34.18_{3.53}$ & $31.94_{2.75}$ & $36.53_{3.80}$ & $51.38_{3.94}$ & $15.12_{1.66}$ & $30.22_{2.54}$ & $25.17_{1.49}$ & $29.65_{2.92}$ & $30.92_{0.70}$ \\
Qwen3-VL-8B-Thinking & $28.27_{2.12}$ & $49.55_{3.61}$ & $11.21_{1.17}$ & $33.92_{2.70}$ & $42.07_{1.78}$ & $29.73_{3.69}$ & $26.70_{2.37}$ & $32.53_{3.87}$ & $47.83_{3.91}$ & $14.10_{1.70}$ & $24.90_{2.30}$ & $28.75_{1.59}$ & $24.20_{2.80}$ & $28.01_{0.67}$ \\
Qwen3-VL-8B-Instruct & $25.31_{2.07}$ & $45.65_{3.88}$ & $13.61_{1.52}$ & $29.97_{2.99}$ & $34.85_{1.70}$ & $27.46_{2.95}$ & $25.27_{2.49}$ & $27.98_{3.47}$ & $35.94_{3.41}$ & $10.66_{1.43}$ & $24.41_{2.60}$ & $20.40_{1.31}$ & $25.07_{2.77}$ & $24.51_{0.64}$ \\
Qwen3-VL-4B-Thinking & $24.23_{2.08}$ & $45.07_{3.71}$ & $12.83_{1.35}$ & $30.66_{2.86}$ & $38.53_{1.75}$ & $29.72_{3.39}$ & $24.89_{2.35}$ & $31.29_{3.31}$ & $46.69_{4.21}$ & $14.92_{1.89}$ & $23.85_{2.42}$ & $25.31_{1.39}$ & $22.60_{2.54}$ & $26.18_{0.65}$ \\
Qwen3-VL-4B-Instruct & $20.23_{1.91}$ & $21.00_{2.84}$ & $9.94_{1.32}$ & $31.04_{3.02}$ & $23.02_{1.35}$ & $21.62_{3.07}$ & $18.57_{1.91}$ & $21.35_{2.85}$ & $35.00_{3.56}$ & $7.69_{1.26}$ & $18.40_{1.97}$ & $11.23_{1.10}$ & $20.83_{2.73}$ & $18.05_{0.56}$ \\
Qwen3-VL-2B-Thinking & $11.81_{1.33}$ & $24.58_{3.13}$ & $5.43_{0.97}$ & $19.67_{2.24}$ & $17.53_{1.39}$ & $13.32_{2.18}$ & $13.58_{1.47}$ & $16.09_{2.89}$ & $26.97_{3.27}$ & $9.03_{1.38}$ & $17.61_{2.22}$ & $13.81_{1.06}$ & $12.36_{1.88}$ & $13.87_{0.47}$ \\
Qwen3-VL-2B-Instruct & $11.13_{1.53}$ & $19.71_{3.07}$ & $5.28_{0.86}$ & $19.82_{2.22}$ & $17.43_{1.27}$ & $12.21_{1.95}$ & $9.17_{1.33}$ & $14.72_{2.62}$ & $12.77_{2.33}$ & $6.43_{1.00}$ & $12.83_{1.96}$ & $5.88_{0.66}$ & $14.04_{2.12}$ & $11.15_{0.43}$ \\
\midrule
\multicolumn{15}{l}{\quad\textit{Other Open-source}} \\
Skywork-R1V3-38B  & $27.12_{0.74}$ & $47.94_{2.92}$ & $15.30_{1.63}$ & $32.37_{2.44}$ & $36.84_{0.69}$ & $37.25_{1.80}$ & $26.43_{2.63}$ & $28.27_{1.95}$ & $41.71_{4.53}$ & $14.76_{1.96}$ & $30.10_{2.73}$ & $27.38_{0.26}$ & $26.42_{0.26}$ & $27.76_{0.58}$ \\
\midrule
\multicolumn{15}{l}{\textit{Korean-specialized Models}} \\
VARCO-VISION-2.0-14B & $11.90_{0.79}$ & $34.76_{4.78}$ & $7.94_{0.85}$ & $17.83_{2.30}$ & $22.03_{2.71}$ & $23.46_{3.16}$ & $21.89_{1.09}$ & $14.05_{2.80}$ & $12.68_{1.90}$ & $7.80_{2.64}$ & $18.84_{1.75}$ & $14.97_{0.57}$ & $13.31_{1.31}$ & $15.55_{0.50}$ \\
HyperCLOVA-3B & $8.42_{0.98}$ & $29.74_{2.98}$ & $6.33_{0.49}$ & $15.17_{1.40}$ & $18.54_{0.41}$ & $15.80_{2.14}$ & $13.38_{0.67}$ & $13.43_{3.83}$ & $9.86_{2.44}$ & $6.16_{0.70}$ & $14.20_{1.75}$ & $16.21_{0.93}$ & $9.53_{1.86}$ & $12.66_{0.18}$ \\
VARCO-VISION-2.0-1.7B & $8.09_{1.21}$ & $21.34_{1.50}$ & $5.95_{2.38}$ & $16.07_{0.96}$ & $17.79_{0.63}$ & $16.22_{1.16}$ & $12.70_{0.32}$ & $12.88_{1.08}$ & $12.54_{5.35}$ & $8.11_{0.68}$ & $12.81_{1.01}$ & $12.13_{0.72}$ & $10.46_{3.57}$ & $11.87_{0.46}$ \\
\bottomrule
\end{tabular}
\caption{Complete performance across all 13 categories for all evaluated models (scores in \%). All scores are reported as mean$_{SE}$, where SE is the standard error over 3 independent runs (n=3).}
\label{tab:detailed_all_categories}
\end{table*}
\end{landscape}

\section{Additional Results \& Analysis}
\label{app:scale_domain}

\subsection{Full Results}
\label{app:full_results}

Table~\ref{tab:detailed_all_categories} reports the full category-wise results for the 45 evaluated models; we will continuously update the leaderboard with newly released models.

\begin{figure}[t]
  \centering
  \begin{subfigure}[t]{0.49\columnwidth}
    \centering
    \includegraphics[width=\linewidth]{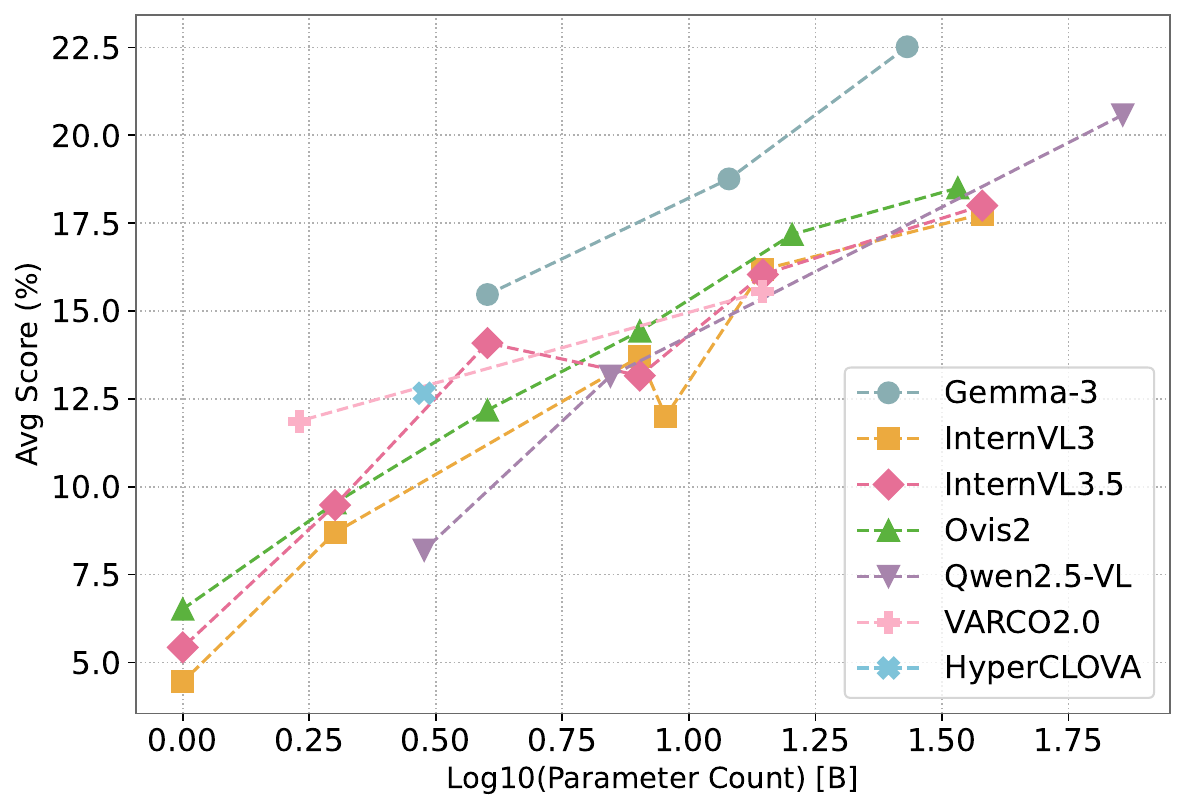}
    \caption{Performance scaling with model size. Accuracy rises up to $\sim$10B parameters but improves more slowly thereafter.}
    \label{fig:output_plot}
  \end{subfigure}\hfill
  \begin{subfigure}[t]{0.49\columnwidth}
    \centering
    \includegraphics[width=\linewidth]{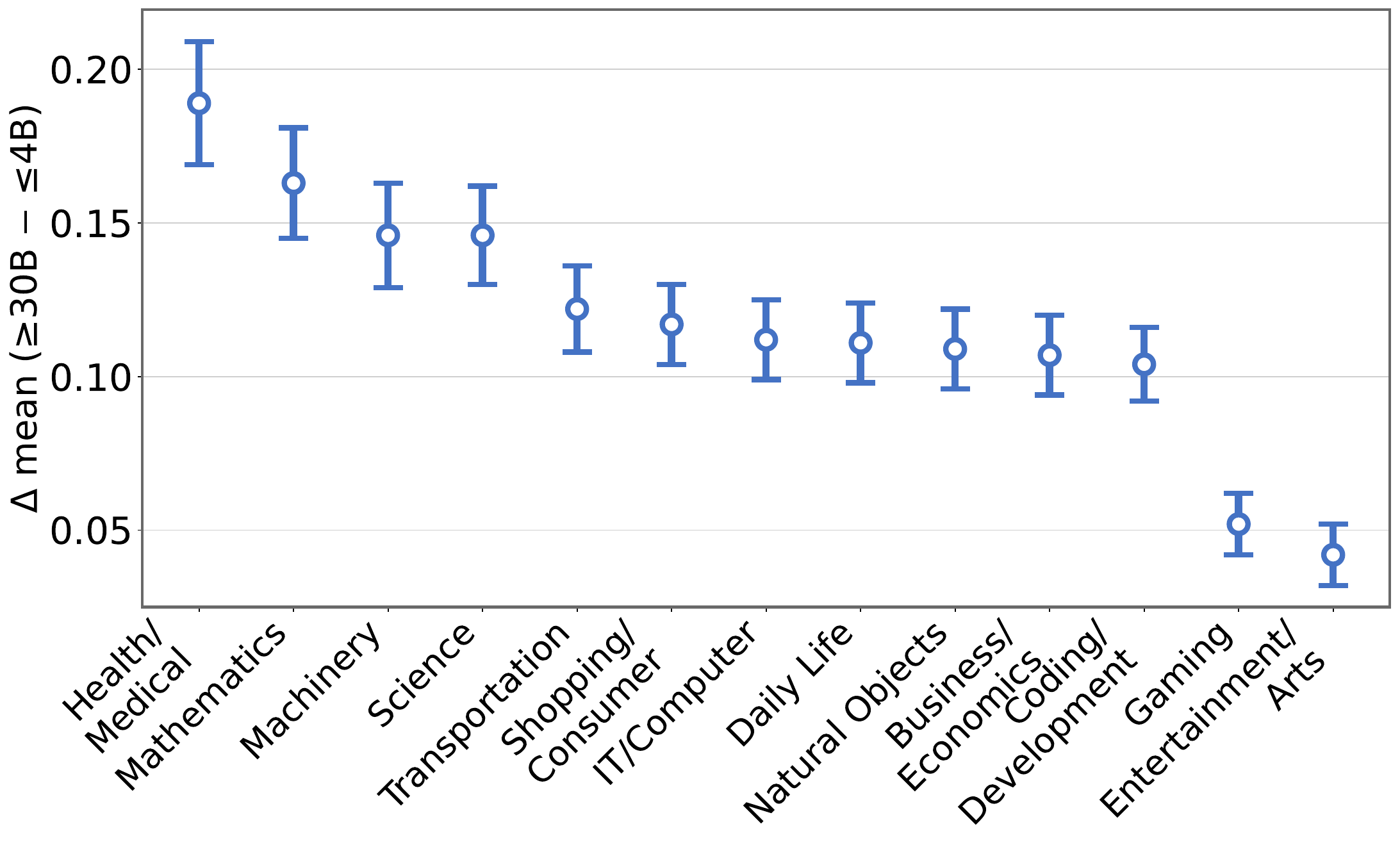}
    \caption{Domain-level results. Health/Medical yields the highest accuracy, whereas Entertainment/Gaming remains the most challenging.}
    \label{fig:domain}
  \end{subfigure}
  \caption{Scaling and domain-level performance on \dataset.}
  \label{fig:scaling_domain}
\end{figure}

\subsection{Performance by Model Scale}

Grouping models by size tiers (Small $\leq$4B, Medium 8–14B, Large $\geq$30B) reveals a clear scaling trend: performance improves with size. Large models reach a mean score of 0.3009 (95\% CI [0.2974, 0.3046]), more than double Medium (0.1460) and triple Small (0.0854). All pairwise differences are significant (permutation $p\!\approx\!0.001$) with large effect sizes (Large–Small $\Delta=+0.2155$, $d\!\approx\!0.78$), confirming that scaling reliably enhances multimodal reasoning.

However, gains become less pronounced beyond about 10B parameters. Accuracy still rises but with smaller marginal improvements (Figure~\ref{fig:output_plot}), indicating that scale alone cannot close the gap. Further progress likely depends on advances in reasoning and cultural grounding.

At the family level, commercial systems (Gemini, GPT, Sonar) consistently outperform open-weight models (e.g., InternVL3), with effect sizes around $d=0.7$–1.2 (e.g., Gemini-2.5-Pro vs InternVL3 $\Delta\!\approx\!0.49$, $d\!\approx\!1.21$). Thus, both scaling and architectural or cultural factors jointly drive performance.

\subsection{Performance by Domain}

Performance varies widely across the 13 domains (global mean = 0.1987, 
range 0.1179–0.332). Health/Medical achieves the highest checklist 
satisfaction (0.332), followed by Science (0.250), while 
Entertainment/Arts (0.118) and Gaming (0.119) remain the most 
challenging. 

Within all domains, large models ($\geq$30B) consistently 
outperform small models ($\leq$4B) (permutation $p<0.05$), with 
the largest gains in Health/Medical ($\Delta=+0.189$) and 
Mathematics ($\Delta=+0.163$). Even in Gaming and Entertainment, 
scale effects remain positive though absolute performance stays low 
(Figure~\ref{fig:domain}).

\subsection{Investigating Failure Modes}
\label{app:InvestigatingFailureModes}

In Table~\ref{tab:detailed_all_categories}, we observe that VARCO-VISION and HyperCLOVA X—two Korean-focused VLMs—underperform multilingual counterparts of similar scale. 
While the precise reasons remain unclear due to the closed nature of these models and limited information about their training, 
we propose two possible explanations:

\begin{enumerate}[label=(\Alph*)]
\item \textbf{Training Data Coverage.} Current benchmarks that capture progress on culturally grounded, information-deficient queries are scarce. 
Model developers may not have explicitly emphasized such aspects in their training data, leading to weaker performance on this type of evaluation.
\item \textbf{Pretraining Scale and Robustness.} Robustness to imperfect or fragmented user queries may emerge from exposure to large-scale, diverse pretraining corpora. 
Larger multilingual models are more likely to encounter noisy, colloquial, or partially specified inputs, thereby preparing them better for benchmarks of this kind.
\end{enumerate}

\subsection{Visual-Text Grounding Error Analysis}
\label{app:vtg_unmasking}

Table~\ref{tab:error_shift} shows that visual-text grounding (VTG) errors increase from 5.2\% to 16.6\% after explicitation. To understand whether this reflects newly introduced errors or pre-existing failures being reclassified, we tracked individual (question, model) pairs across conditions.

\paragraph{Error Tracking.}
Of the 461 VTG errors in the explicitated condition, 401 (87.0\%) were \textit{newly surfaced}---cases that were already errors under original queries but classified under different failure categories. Only 60 (13.0\%) were \textit{persistent} VTG errors present in both conditions. Additionally, 104 VTG errors from the original condition were resolved by explicitation.

\paragraph{Source of Newly Surfaced VTG Errors.}
The 401 newly surfaced cases were previously classified under the following failure categories in the original condition (multi-label; percentages sum to $>$100\%):

\begin{table}[h]
\centering
\begin{tabular}{lc}
\toprule
\textbf{Original Failure Category} & \textbf{\%} \\
\midrule
Lack of explicitness & 70.8 \\
Object recognition & 55.6 \\
Procedural reasoning & 29.2 \\
Cultural concept mismatch & 21.4 \\
\bottomrule
\end{tabular}
\caption{Original failure categories of newly surfaced VTG errors. Most were previously masked by lack-of-explicitness failures.}
\label{tab:vtg_source}
\end{table}

\paragraph{Interpretation.}
This supports an \textit{error unmasking} interpretation: under-specified queries produce vague responses that fail for surface-level reasons. Once explicitation removes this ambiguity, models are forced to engage with specific visual regions, exposing deeper grounding failures previously masked by the dominant lack-of-explicitness error mode. Supporting this, VTG error severity remains virtually identical across conditions (severe: 84.8\% $\to$ 83.7\%), indicating that explicitation reveals pre-existing failures rather than creating new ones.

\paragraph{Example.}
In one case (idx=22), the model's response to the original under-specified query was vague and non-committal, merely identifying the character's reading while omitting details---annotated as ``lack of explicitness.'' When given the explicitated query, the model attempted a specific answer but misread the character \begin{CJK}{UTF8}{mj}惹\end{CJK} as \begin{CJK}{UTF8}{mj}芯\end{CJK}---annotated as ``visual-text grounding.'' The same question produced different error types because explicitation forced the model to engage with the specific visual region, shifting the failure from surface-level vagueness to a concrete grounding error.

\section{Error Annotation Methodology}
\label{app:error_annotation}

\subsection{Annotation Setup}

We used Claude 3.5 Sonnet as the LLM judge for error annotation, accessed via OpenRouter API with \texttt{temperature=0.0} and \texttt{max\_tokens=2048}. For each error case (model response with score $<$ 1.0), the judge was provided with the original question, gold answer, checklist items, model response, and metadata (source, category, model name, score).

\begin{table*}[t]
\centering
\small

\label{tab:taxonomy}
\begin{tabular}{p{3.6cm}p{9cm}}
\toprule
\multicolumn{2}{l}{\textbf{Root Cause} (select one)} \\
\midrule
language & Misunderstood Korean grammar, negation, particles, or expressions \\
cultural\_knowledge & Lacked Korean-specific cultural/institutional knowledge \\
general\_reasoning & Understood language and context but failed at reasoning \\
\midrule
\multicolumn{2}{l}{\textbf{Failure Category} (select 1--3)} \\
\midrule
object\_recognition & Fails to identify key objects in the image \\
spatial\_reasoning & Misinterprets spatial relations \\
cultural\_concept\_mismatch & Misunderstands Korean-specific concepts or conventions \\
visual\_text\_grounding & Refers to the wrong region/entity relative to the question \\
procedural\_reasoning & Fails to execute multi-step procedures \\
lack\_of\_explicitness & Misses explicit facts demanded by the checklist \\
other & None of the above fit \\
\midrule
\multicolumn{2}{l}{\textbf{Severity}} \\
\midrule
minor & Almost correct; small missing detail \\
moderate & Mixed correctness; partially useful \\
severe & Largely incorrect or misleading \\
\bottomrule
\end{tabular}
\caption{Error annotation taxonomy.}
\end{table*}

\subsection{Annotation Prompt}

\begin{tcolorbox}[colback=gray!5, colframe=gray!50, boxrule=0.5pt, arc=2pt, fontupper=\small]
\textbf{System prompt:}

You are an impartial error analysis assistant for a Korean multimodal QA benchmark. Your job is to carefully inspect each example and classify the model's failure according to a predefined taxonomy. Follow the provided schema exactly. Think step by step, but ONLY return the final JSON object in your response. Do NOT include explanations outside the JSON. Be strict and consistent with the taxonomy definitions.

\textbf{User prompt:}

You are given one question-answering example from a Korean multimodal benchmark, together with a model's answer and a detailed checklist used for scoring. Your goal is to analyze WHY the model failed or was only partially correct.

Based on the question, gold answer, checklist, and model answer:

1. Decide the SINGLE most important root cause of failure: ``language'', ``cultural\_knowledge'', or ``general\_reasoning''

2. Choose 1--3 failure\_categories describing HOW the error manifests

3. Choose severity: ``minor'', ``moderate'', or ``severe''

4. Provide analysis\_comment: 2--3 sentences in Korean explaining why the answer is wrong or incomplete

[Output format]

Return ONLY a single JSON object:

\{``root\_cause'': ``...'', ``failure\_categories'': [...], ``severity'': ``...'', ``analysis\_comment'': ``...''\}

[Metadata]

- source: \{source\}
- category: \{category\}
- question\_idx: \{question\_idx\}
- model\_name: \{model\}
- model\_score: \{score\}

[Question] \{question\}

[Gold answer] \{answer\}

[Checklist items] \{checklist\}

[Model answer] \{model\_response\}
\end{tcolorbox}



\end{document}